%% file: main.tex
\documentclass[10pt,twocolumn,letterpaper]{article}

\usepackage{cvpr}      
\input{preamble}
\definecolor{cvprblue}{rgb}{0.21,0.49,0.74}
\usepackage[dvipsnames]{xcolor}
\usepackage[pagebackref,breaklinks,colorlinks,allcolors=cvprblue]{hyperref}
\usepackage{array}
\usepackage{multirow}
\usepackage{multicol}
\usepackage{nicematrix}
\usepackage{diagbox} 
\usepackage{algorithm}
\usepackage{algorithmic}
\usepackage{xcolor} 
\usepackage{listings} 
\usepackage{lipsum}
\usepackage{amsmath}
\usepackage{colortbl}
\usepackage{tabularray}
\usepackage{empheq}
\usepackage{amsmath, amssymb, amsfonts} 
\usepackage{mathtools} 
\usepackage{mathrsfs} 
\usepackage{dutchcal}

\def\paperID{*****} 
\def\confName{CVPR}
\def\confYear{2026}

\newcommand{\emoji}{\includegraphics[height=2.5\fontcharht\font`\B]{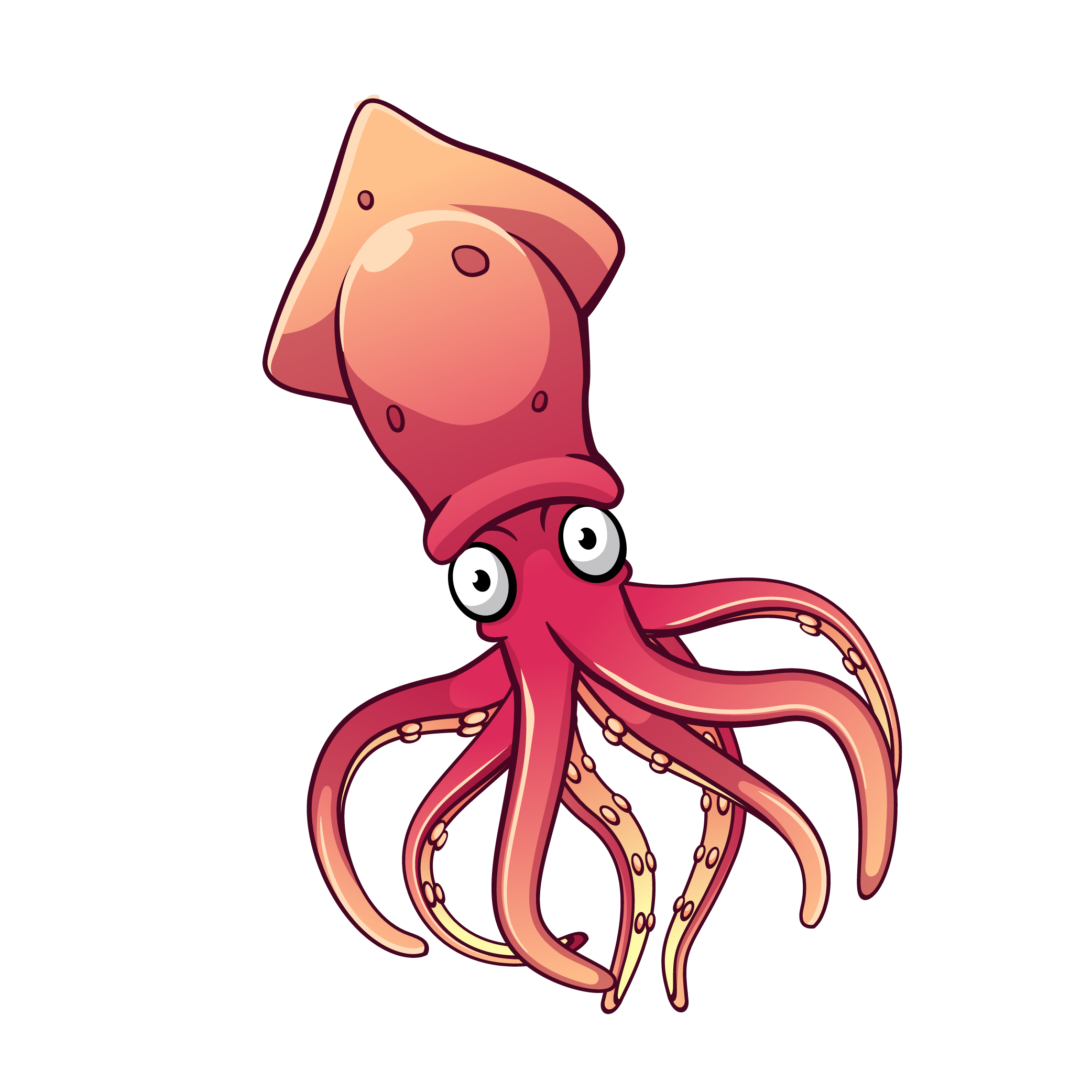}}

\title{
\raisebox{-1ex}{\emoji}{Octopus}: History-Free Gradient Orthogonalization for Continual Learning in Multimodal Large Language Models
}

\author{
    Yuehao Liu$^{1}$ \quad Shanyan Guan$^{2}$ \quad Weijia Zhang$^{1}$ \\
    \quad Xuanming Shang$^{1}$ \quad Yanhao Ge$^{2}$ \quad Wei Li$^{2}$ \quad Chao Ma$^{1}$\thanks{~Corresponding author.} \\
    ${}^{1}$ MoE Key Lab of Artificial Intelligence, AI Institute, Shanghai Jiao Tong University \\
    ${}^{2}$ vivo Mobile Communication Co., Ltd.\\
    {\tt\small \{yuehao.liu, weijia.zhang, sxm2021, chaoma\}@sjtu.edu.cn} \\
    {\tt\small \{guanshanyan, halege, liwei.yxgh\}@vivo.com} \\
    {\small Project page: \url{https://fxmangd26.github.io/Octopus/}}
    }

\usepackage{amssymb, bbding}
\usepackage[accsupp]{axessibility}  

\newcommand{\our}{Octopus}
\newcommand{\yes}{\textcolor{red!80!black}\Checkmark}
\newcommand{\no}{\textcolor{green!70!black}\XSolidBrush}

\usepackage{enumitem,amssymb, pifont}
\newlist{todolist}{itemize}{2}
\setlist[todolist]{label=$\square$}

\usepackage{CJKutf8}
\usepackage[dvipsnames]{xcolor}

\begin{document}


\maketitle


\input{sec/0_abstract}    
\input{sec/1_intro}

\input{sec/2_related_work}

\input{sec/3_method}
\input{sec/4_experiments}
\input{sec/5_conclusion}
\input{sec/6_acknowledgement.tex}

{
    \small
    \bibliographystyle{ieeenat_fullname}
    \bibliography{main}
}

\input{sec/X_suppl}


\end{document}

%% file: sec/0_abstract.tex
\begin{abstract}
Continual learning in multimodal large language models (MLLMs) aims to sequentially acquire knowledge while mitigating catastrophic forgetting, yet existing methods face inherent limitations: architecture-based approaches incur additional computational overhead and often generalize poorly to new tasks, rehearsal-based methods rely on storing historical data, raising privacy and storage concerns, and conventional regularization-based strategies alone are insufficient to fully prevent parameter interference. We propose \our, a two-stage continual learning framework based on History-Free Gradient Orthogonalization (HiFGO), which enforces gradient-level orthogonality without historical task data. Our proposed two-stage finetuning strategy decouples task adaptation from regularization, achieving a principled balance between plasticity and stability. Experiments on UCIT~\cite{guo2025hide} show that \our~establishes state-of-the-art performance, surpassing prior SOTA by 2.14\% and 6.82\% in terms of \textit{Avg} and \textit{Last}.
\end{abstract}

%% file: sec/1_intro.tex
\vspace{-0.5cm}
\section{\textcolor{black}{Introduction}}
\label{fig_radar}
Continual learning~\cite{parisi2019continual, rolnick2019experience, wang2022learning, zhou2024class, cao2024continual, srivastava2024improving, wang2024comprehensive}, sequentially learning across multiple task without forgetting, 
allows multimodal large language models (MLLMs)~\cite{liu2024improved, yin2024survey} to incrementally integrate knowledge across tasks, thereby exhibiting human-like adaptability when encountering novel scenarios. Catastrophic forgetting~\cite{mccloskey1989catastrophic, french1999catastrophic, ahn2021ss, douillard2020podnet, hu2021distilling, wu2019large, wang2024meet} constitutes the fundamental challenge in continual learning, referring to the degradation of previously acquired knowledge as a model adapts to new tasks.

\begin{figure}
    \centering
    \includegraphics[width=\linewidth]{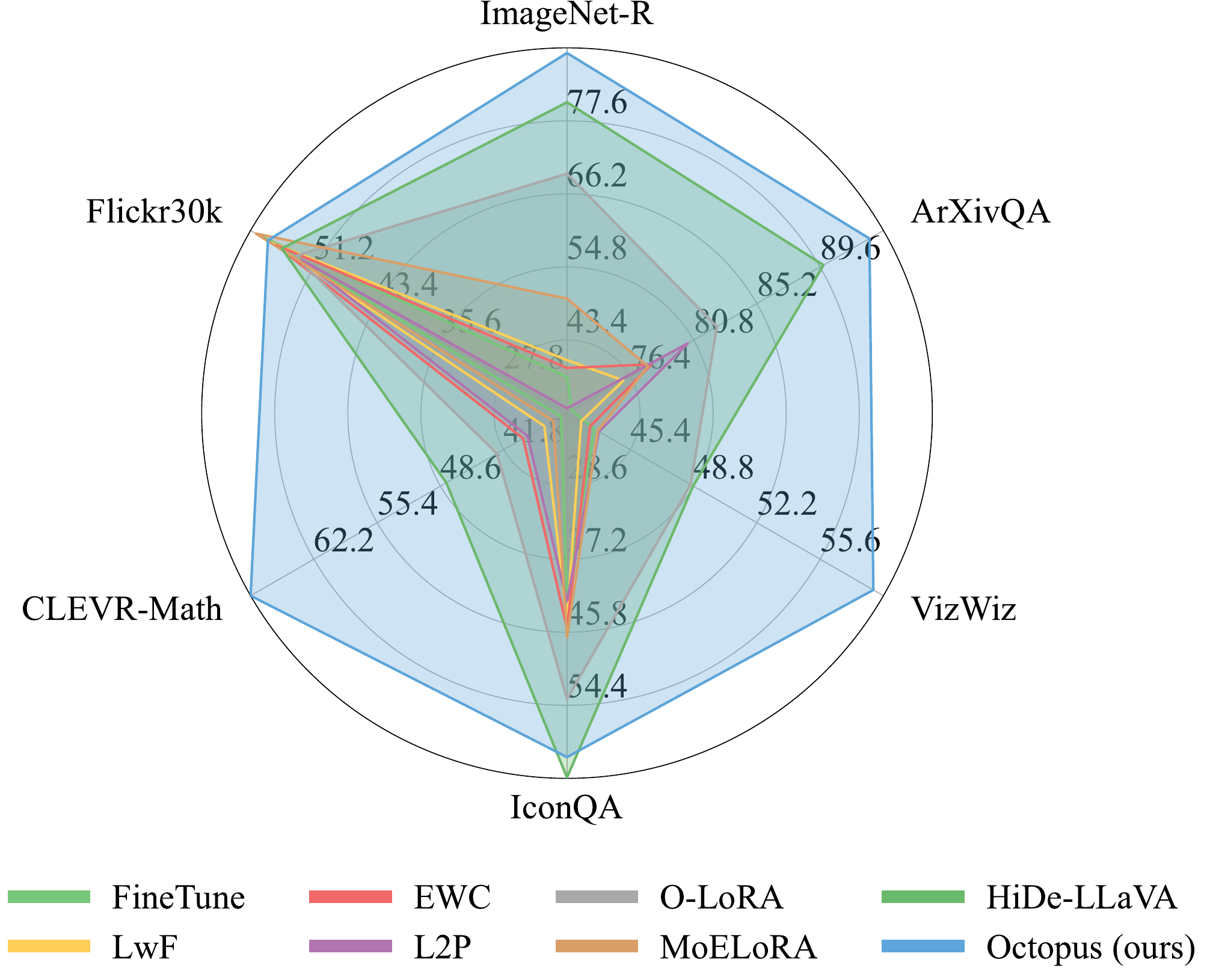}
    \caption{Performance comparison between Octopus (ours) and existing approaches on UCIT~\cite{guo2025hide} in terms of \textit{Last}. Results demonstrate that Octopus establishes a new SOTA performance and outperforms all competing methods by a substantial margin.}
    \vspace{-0.15cm}
\end{figure}

Current continual learning approaches for MLLMs can be broadly classified into three categories: \textit{architecture-based}, \textit{rehearsal-based} and \textit{regularization-based} methods. 
Architecture-based methods~\cite{chen2024coin, guo2025hide, huai2025cl} typically assign several LoRA modules to store task-specific information, which tends to deteriorate the model’s generalization to unseen tasks and sacrificing computational efficiency during inference. 
Rehearsal-based methods~\cite{smith2024adaptive, chaudhry2019continual, smith2024adaptive} maintain memory modules that store historical task information, such as past data or intermediate layer activations. However, in real-world applications, access to historical task data may be impractical or raise data privacy concerns, while maintaining replay buffers incurs additional storage overhead.
In contrast, regularization-based approaches~\cite{wang2023orthogonal, zhu2025bilora} are not subject to the aforementioned limitations. These methods seek to constrain parameter updates within subspaces that minimally affect previously tasks to mitigate interference.

\begin{figure*}
    \centering
    \includegraphics[width=\textwidth]{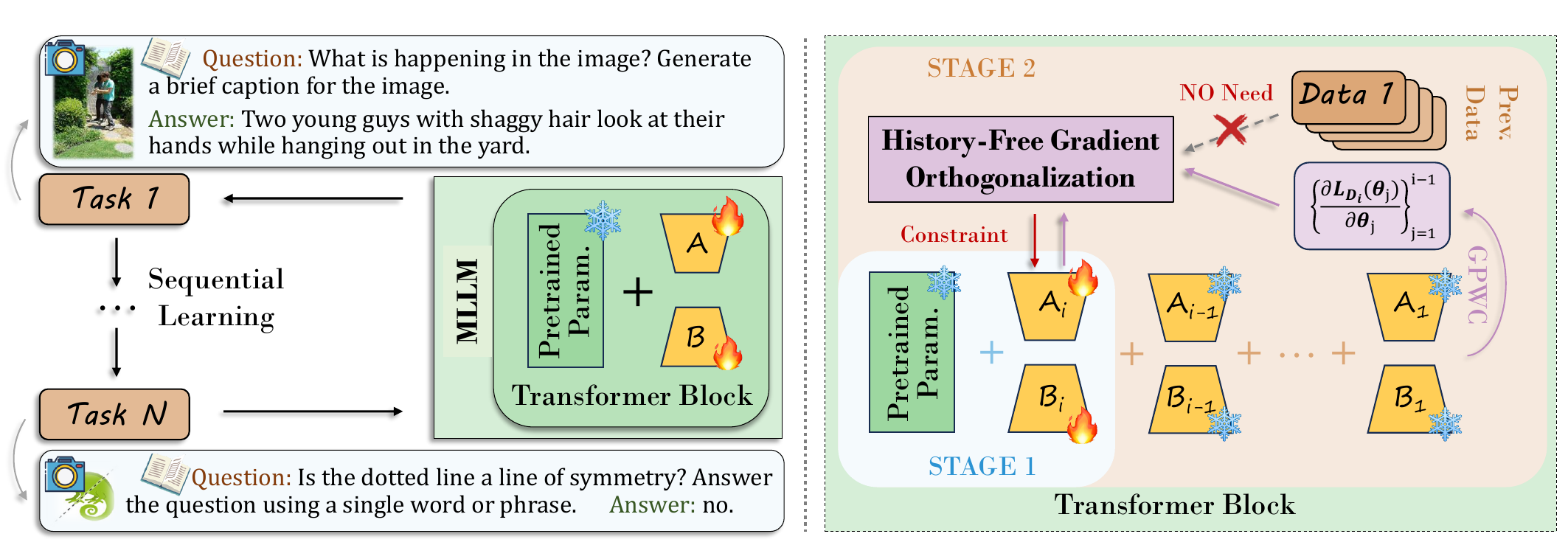}
    \caption{Overall pipeline for continual learning in MLLMs (\textit{left}) and architecture of our proposed \our\ (\textit{right}). In the context of continual learning, MLLMs are required to sequentially learn multiple tasks while overcoming the challenge of catastrophic forgetting caused by non-stationary data distributions. To address this, we propose \our, which adopts a two-stage fine-tuning paradigm. In the first stage, MLLM learns task-specific knowledge without constraints, enabling full adaptation to current task. In the second stage, we apply History-Free Gradient Orthogonalization (HiFGO) to mitigate parameter interference, while simultaneously constraining the parameter updates within an optimal solution space, thereby maintaining a effective balanced trade-off between plasticity and stability.}
    \label{fig_main}
    \vspace{-0.2cm}
\end{figure*}

The primary objective of regularization is to mitigate parameter interference, wherein the acquisition of parameters for new tasks does not compromise the performance of previously learned tasks, thereby alleviating catastrophic forgetting.
primarily focus on enforcing parameter orthogonality.~\cite{wang2023orthogonal, zhu2025bilora} primarily focus on enforcing parameter orthogonality. However, studies in model merging~\cite{stoica2024model} suggest that parameter orthogonality is insufficient to fully prevent parameter interference. 
We theoretically demonstrate that, beyond parameter orthogonality, it is of greater significance that gradient orthogonality must be enforced, which is consistent with prior works~\cite{farajtabar2020orthogonal, yang2023restricted}; however, existing gradient orthogonality-based methods~\cite{farajtabar2020orthogonal, yang2023restricted} remain fundamentally constrained by their reliance on historical data.

To address this limitation, we propose \our, a two-stage continual learning framework based on history-free gradient orthogonalization. Specifically, we propose History-Free Gradient Orthogonalization (HiFGO) characterizing sensitivity of previous task parameters within current data distribution, leveraging only past weights (instead of past data) and current task data to enforce orthogonal constraints in gradient space. 
Moreover, we observed in experiments that the regularization constraints tend to compete with the objectives of original task, which is consistent with studies in multi-task learning~\cite{zhang2018overview, caruana1997multitask, crawshaw2020multi, zhang2021survey}. To alleviate this competition, we propose a two-stage finetuning strategy maintaining parameters in proximity to the optimal solution under incorporation of regularization constraints. 

Extensive experiments on UCIT~\cite{guo2025hide} benchmark demonstrate that our framework achieves state-of-the-art (SOTA), surpassing the previous SOTA~\cite{guo2025hide} by 2.14\% and 6.82\% in terms of \textit{Avg} and \textit{Last}, respectively. It indicates that our proposed HiFGO effectively preserves previously learned knowledge while learning new tasks. Moreover, our two-stage finetuning strategy substantially enhances the performance ceiling of regularization-based methods, allowing them to approach even exceed the performance of multi-task training while preserving the efficacy of regularization, achieving effective balance between plasticity and stability.

%% file: sec/2_related_work.tex
\section{\textcolor{black}{Related Work}}
\label{sec:related_work}

\paragraph{\textcolor{black}{Parameter-Efficient Model Adaptation.}}
MLLMs have achieved remarkable success across diverse tasks~\cite{glm2024chatglm, dubey2024llama, bai2025qwen2, zhao2025chartedit, lu2023mathvista, radford2019language}; however, maintaining a separately fine-tuned model for each new task incurs prohibitive computation and storage costs. Parameter-efficient fine-tuning (PEFT)~\cite{houlsby2019parameter, chen2022adaptformer, lester2021power, li2021prefix, jia2022visual, ran2025correlated} have been proposed to address this issue. Approaches such as Adapters~\cite{houlsby2019parameter} and AdaptFormer~\cite{chen2022adaptformer} introduce trainable modules into pretrained networks, yet they increase model complexity and fail to ensure task isolation. Prompt-tuning~\cite{lester2021power} and Prefix-tuning~\cite{li2021prefix} introduce learnable representations into  into Transformer layers, but their expressiveness is limited to input-level manipulation. Besides, LoRA and its variants decompose weight updates into low-rank subspace, enabling efficient and scalable adaptation, and thus have become mainstream in MLLM fine-tuning, and have been certified to exhibit a lower degree of forgetting compared to full-parameter fine-tuning. 

\vspace{-15pt}
\paragraph{\textcolor{black}{Conventional Continual Learning.}}
The central objective of continual learning is to alleviate catastrophic forgetting in sequential task learning. Classical approaches such as EWC~\cite{kirkpatrick2017overcoming} and LwF~\cite{li2017learning} address this by introducing regularization terms that penalize updates of parameters critical to previous tasks. Methods such as DER~\cite{buzzega2020dark}, DGR~\cite{shin2017continual} and GEM~\cite{lopez2017gradient} implicitly preserve knowledge from previous tasks in order to mitigate forgetting by reinforcing historical knowledge during training of new tasks. Another line of research such as Piggyback~\cite{mallya2018piggyback} tackles catastrophic forgetting through architectural adaptation, expanding base model with task-specific modules or adaptive parameters to enable sequential learning without forgetting.

\vspace{-10pt}
\paragraph{\textcolor{black}{Continual Learning for Multi-Modal Language Models.}}
Prompt-based methods such as L2P~\cite{wang2022learning}, DualPrompt~\cite{wang2022dualprompt}, and CODA-Prompt~\cite{smith2023coda} enhance knowledge retention by tuning soft prompt vectors without altering model weights. MoE-based approaches, including HiDe-LLaVA~\cite{guo2025hide} and MoILE~\cite{jia2025hierarchical}, assign task-specific experts via routing mechanisms. While effective, both incur additional inference or storage costs.
Orthogonal gradient strategies like OGD~\cite{farajtabar2020orthogonal} mitigate task interference by constraining updates to directions orthogonal to previous gradients, yet require access to past task data. Recent extensions—OLoRA~\cite{wang2023orthogonal}, BiLoRA~\cite{zhu2025bilora}, and InfLoRA~\cite{liang2024inflora}—replace stored gradients with parameter vectors but still face limited efficacy and the inherent plasticity–stability trade-off.
\our{} alleviates these limitations, achieving inference-efficient, rehearsal-free, and highly effective solution to continual learning.

%% file: sec/3_method.tex
\section{\textcolor{black}{Preliminaries}}
\begin{figure}
    \centering
    \includegraphics[width=\linewidth]{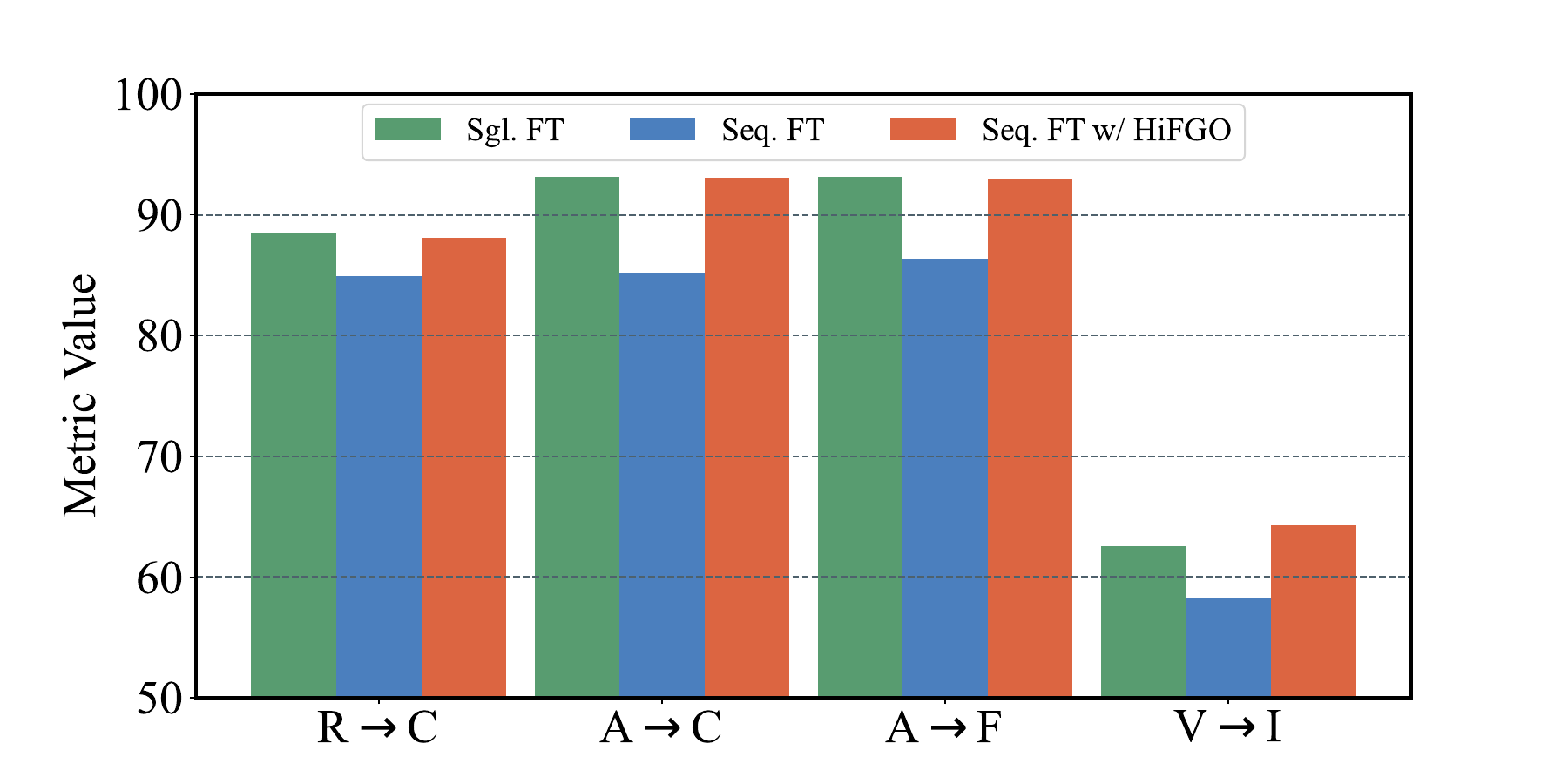}
    \caption{Effectiveness of HiFGO. We present the performance of single-task finetuning, sequential fine-tuning of two tasks, and fine-tuning with HiFGO constraints added after sequential fine-tuning (higher is better). The abbreviations in the table represent dataset names, with details provided in Sec. \ref{ablation_on_task_order}.}
    \label{fig_method}
    \vspace{-0.1cm}
\end{figure}
\subsection{\textcolor{black}{Low-Rank Adaptation}}
Low-Rank Adaptation (LoRA)~\cite{hu2022lora} introduces a parameter-efficient finetuning paradigm for large-scale pre-trained models.
Formally, consider a pre-trained parameter matrix $W_0 \in \mathbb{R}^{d \times k}$. Instead of optimizing $W_0$, LoRA parameterizes the weight update as product of two low-rank matrices:
\begin{equation}
\Delta W = B A,
\end{equation}
where $A \in \mathbb{R}^{r \times k}$ and $B \in \mathbb{R}^{d \times r}$ with $r \ll \min(d, k)$. The adapted weight is then defined as:
\begin{equation}
W = W_0 + \Delta W = W_0 + B A.
\end{equation}

During fine-tuning, only the low-rank factors $A$ and $B$ are updated while original weights $W_0$ remain frozen. 
Owing to its efficiency and scalability, LoRA has become a popular technique for PEFT in various domains and and has been widely applied in continual learning tasks.

\subsection{\textcolor{black}{Continual Learning}}
Continual Learning (CL) considers the problem of incrementally acquiring knowledge from a sequence of tasks $T_1, T_2, \dots, T_N$ while preserving robust performance on previous tasks. Let $f_\theta$ denote a model parameterized by $\theta$, and $\mathcal{D}_i = {(x_k, y_k)}$ denote the dataset corresponding to task $T_i$. From a probabilistic perspective, the objective of CL can be formulated as maximizing the expected predictive likelihood across the entire task sequence:
\begin{equation}
\theta^* = \arg \max_\theta \frac{1}{N} \sum_{i=1}^N \frac{1}{|\mathcal{D}_i|} \sum_{(x_k, y_k) \in \mathcal{D}_i} \log p_\theta(y_k \mid x_k),
\end{equation}
where $p_\theta(y_k \mid x_k)$ represents conditional probability of observing label $y_k$ given input $x_k$ under the model $f_\theta$. 

This suggests that the goal of continual learning is to learn a unified parameter set that is generalized across all datasets, inherently balancing plasticity and stability during training process. However, in practice, it is fundamentally constrained by catastrophic forgetting (CF)~\cite{mccloskey1989catastrophic, french1999catastrophic}, a phenomenon in which sequential optimization on new tasks induces rapid deterioration of previously acquired knowledge.

\section{\textcolor{black}{Methodology}}
\label{sec:method}

In this work, we propose \our, a two-stage continual learning framework based on history-free gradient orthogonalization for MLLMs. We first provide a theoretical justification in Sec.~\ref{theoretical_analysis} for why the gradient directions of previous tasks intrinsically capture model’s sensitivity subspace, thereby suggesting that beyond parameter orthogonality, it is of greater significance that gradient orthogonality must be enforced to mitigate parameter interference.
Building upon this insight, we introduce a history-free gradient orthogonalization in Sec. ~\ref{gradient_orthogonal} to mitigate parameter interference, which operates without reliance on historical task data. Our method offers a effective and data-efficient solution to mitigating catastrophic forgetting in continual learning. 

Moreover, to mitigate the degradation in fine-tuning performance induced by parameter regularization, we introduce a two-stage finetuning framework in Sec.~\ref{two_stage_training} that allows LoRA to satisfy the imposed constraints while remaining close to optimal manifold. This design effectively balances plasticity and stability, thereby further enhancing the model’s capability for continual learning. The overall architecture of our proposed method is illustrated in Fig.~\ref{fig_main}.

\subsection{\textcolor{black}{
Analysis on Parameter Interference}}
\label{theoretical_analysis}
We denote $W_0$ as the pretrained parameters of MLLM, $\theta_i$ as the LoRA parameters after fine-tuning on Task~$i$, $\mathcal{D}_i$ as the corresponding training dataset for Task~$i$, and $L_{\mathcal{D}_i}{(\theta)}$ represents the expectation of loss function evaluated on $\mathcal{D}_i$ given model parameters $\theta$. 
For clarity, we first consider a sequential learning scenario involving two tasks, which can be naturally generalized to the multi-task case.

Let $W_0$ denotes the pretrained weights, $\theta_1$ and $\theta_2$ as the optimized LoRA weights of Task~$1$ and Task~$2$, respectively.  
Our objective is to ensure that the training on Task~$2$ does not degrade the performance achieved on Task~$1$. Formally, by defining $\theta_1' = W_0 + \theta_1$, we require the following condition to hold, which we define as the \textbf{lossless condition}:
\begin{equation}
    L_{\mathcal{D}_1}{(\theta_1')} = L_{\mathcal{D}_1}{(\theta_1' + \theta_2)}.
\end{equation}

We perform Taylor expansion around $\theta_1'$ as:
\begin{equation}
L_{\mathcal{D}_1}(\theta_1' + \theta_2) = 
L_{\mathcal{D}_1}(\theta_1') + \langle \frac{\partial L_{\mathcal{D}_1}(\theta_1')}{\partial\theta_1'}, \theta_2 \rangle + \mathcal{O}(||\theta_2||^2),
\label{eq:taylor}
\end{equation}
where $\mathcal{O}(||\theta_2||^2)$ denotes the second- and higher-order residual terms.  
Since LoRA weights are typically much smaller in magnitude than pretrained weights during fine-tuning, $\mathcal{O}(||\theta_2||^2)$ can be safely neglected. Therefore, satisfying the lossless condition reduces to:
\begin{equation}
\langle \frac{\partial L_{\mathcal{D}_1}(\theta_1')}{\partial\theta_1'}, \theta_2 \rangle = 0.
\label{eq:orth}
\end{equation}

This observation suggests that the orthogonality between parameters of Task~$2$ and gradients of Task~$1$ on $\mathcal{D}_1$ provides a principled guarantee to effectively mitigate parameter interference. In contrast, parameter orthogonality—\ie, OLoRA~\cite{wang2023orthogonal}—fails to guarantee lossless disentanglement, as $\frac{\partial L_{\mathcal{D}_1}(\theta_1')}{\partial\theta_1'}$ and $\theta_1$ may correspond to distinct directional semantics. Specifically, derived through fine-tuning from pretrained weights, $\theta_1$ represents the trajectory from pretrained weights toward local optimum, rather than the instantaneous gradient direction at $\theta_1'$. Furthermore, the update direction of $\theta_1$ may vary throughout the optimization process, implying that $\theta_1$ does not necessarily align with a consistent gradient orientation on fixed parameters.

However, in practical scenarios, historical data are often difficult to obtain or restricted due to data privacy concerns. Moreover, as the parameters of previous tasks typically converge to local optima, the gradient at $\theta_1'$ tends to be weak and highly oscillatory, thereby limiting the effectiveness of gradient-based regularization. To address this issue, we propose a history-free gradient orthogonalization paradigm that enables effective parameter disentanglement across tasks.

\subsection{\textcolor{black}{
History-Free Gradient Orthogonalization
}}
\begin{algorithm}[t] 
\renewcommand{\arraystretch}{0.9}
    \caption{Octopus} 
    \label{algo}
    \, \textbf{Input}: Pretrained weights $W_0$; Number of tasks $N$; Data of different tasks $\{\mathcal{D}_i\}_{i=1}^N$
    
    \, \textbf{Output}: Merged LoRA weights $W_0 + \sum\limits_{i=1}^N{\theta_{i, 2}}$

    \, 1: Initialize $\theta_{1, 1}$ following vanilla LoRA

    \, 2: \textbf{for} $i$ in $1:N$ \textbf{do}

    
    \, 3: \qquad Select subset $\mathcal{D}_{i_1}$ and $\mathcal{D}_{i_2}$ from $\mathcal{D}_i$ 

    \, \colorbox{gray!30}{\qquad $\triangledown$ Stage-1 Finetuning \qquad \qquad \qquad \qquad \qquad \qquad}

    \, 4: \qquad Finetune on $\mathcal{D}_{i_1}$ using $\mathcal{L}_1$ (Eq.~\ref{loss_1}) and update $\theta_{i, 1}$

    \, \colorbox{gray!30}{\qquad $\triangledown$ Stage-2 Finetuning \qquad \qquad \qquad \qquad \qquad \qquad}

    \, 5: \qquad Compute all GPWC~(\ref{gpwc}) for Task~$i$

    \, 6: \qquad Finetune on $\mathcal{D}_{i_2}$ using $\mathcal{L}_2$ (Eq.~\ref{loss_2}) and update $\theta_{i, 2}$

    \, 7: \textbf{Return} $W_0 + \sum\limits_{i=1}^N{\theta_{i, 2}}$ as the merged LoRA weight
\end{algorithm}

\label{gradient_orthogonal}
To achieve balance between data privacy preservation and model efficacy, we introduce a novel approach termed \textbf{Hi}story-\textbf{F}ree \textbf{G}radient \textbf{O}rthogonalization (\textbf{HiFGO}). The central objective of this method is to accurately characterize the mutual influence between current and previous tasks without requiring access to any historical task data.

Motivated by SD~\cite{zhao2023rethinking}, which demonstrates that the representation space of each task can be decomposed into a stability-related subspace (task-shared subspace) and a plasticity-related subspace (task-specific subspace), we propose to quantify inter-task interference via \textbf{G}radients of \textbf{P}revious parameters \textbf{W}ithin \textbf{C}urrent data distribution (\textbf{GPWC}\label{gpwc}). Intuitively, GPWC reflects beneficial update directions that encapsulate the reusable knowledge embedded in earlier tasks, effectively capturing the shared representational subspace across tasks. Moreover, since previous parameters have converged to local optima within their respective domain, direction of their gradients on current data distribution also reveals optimization conflicts between tasks, indicating the potential for parameter updates in the current task to degrade performance on earlier ones.

\input{tables/main_result}

To mitigate such interference, we introduce a gradient orthogonality constraint that enforces orthogonality between current parameters and GPWC. This regularization effectively disentangles parameter updates across tasks, promoting stability while maintaining adaptability. Concretely, during fine-tuning of Task~$i$, we incorporate the following orthogonality loss term into the optimization objective:
\begin{equation}
    L_{orth}(\theta_i) = \sum\limits_{j = 1}^{i - 1}\left(\frac{\partial L_{\mathcal{D}_i}(\theta_j')}{\partial\theta_j'}\right)^T\theta_i,
\label{orth-loss}
\end{equation}
where $\theta_j'=W_0+\sum\limits_{m=1}^j{\theta_m}$ is denoted as the merged LoRA weights of Task~$j$, while $\mathcal{L}_{D_i}({\theta_j'})$ is denoted as the loss function of parameter $\theta_j'$ on $\mathcal{D}_i$

We evaluate the efficacy of HiFGO through a controlled sequential fine-tuning protocol. We sequentially finetune an MLLM on two tasks (named Task~$1$ and Task~$2$),which would induce substantial degradation on Task~$1$ in comparison to single-task finetuning. Then we further finetune MLLM on Task~$2$ for a few steps with HiFGO constraint. We evaluate the performance of Task~$1$, as shown in Fig. \ref{fig_method}. The results show that sequential finetuning markedly degrades prior-task performance, whereas introducing HiFGO nearly restores it to the level of single-task finetuning, which demonstrate that HiFGO effectively suppresses parameter interference and preserves previously acquired knowledge.

\vspace{-13pt}
\paragraph{Historical task proxy approximation.}
In Eq. \ref{orth-loss}, the parameters of the current task are required to compute the inner product with the GPWC of each historical task, which causes the training cost to grow linearly with the number of tasks. To address this issue, we introduce a lightweight approximation based on a valid proxy of historical tasks parameters. Instead of computing the constraint with respect to all historical parameters, we use the parameter $\theta_{i-1}'$ from the most recent task as a proxy representing the entire task history, as:
\begin{equation}
L_{orth}'(\theta_i) = \left(\frac{\partial L_{\mathcal{D}_i}(\theta_{i-1}')}{\partial\theta_{i-1}'}\right)^T\theta_i
\label{new_orth}
\end{equation}
This approximation is motivated by the observation that a well-trained continual learning model preserves the performance of earlier tasks after each training stage, thereby implicitly encoding historical knowledge in the latest parameters. As a result, the computational cost of the orthogonal loss is reduced from $O(t)$ to $O(1)$.

\input{tables/Orthogonalization}
\subsection{\textcolor{black}{Two-stage Finetuning Strategy}}
\label{two_stage_training}
Although the magnitudes of LoRA weights are typically much smaller than those of the pretrained parameters during fine-tuning, the vanilla fine-tuning paradigm can still incur non-negligible errors due to the influence of higher-order terms. To alleviate this issue, it is natural to introduce additional regularization terms that constrain the scale of LoRA weights, thereby mitigating the impact of such higher-order effects. However, our experimental observations indicate that imposing such constraint considerably degrades the fine-tuning performance and, consequently, compromises the model’s continual learning capability.

\input{tables/Two-stage_training}

We attribute this phenomenon to two primary factors: (1) the additional constraints substantially reduce the effective parameter search space, and (2) the interference among multiple loss objectives increases the risk of the optimization process being trapped in suboptimal local minima. As a result, the optimized solution can diverge significantly from that obtained through standard fine-tuning.

To address this challenge, we draw inspiration from annealing schedule~\cite{fu2019cyclical} and propose a two-stage finetuning strategy, which enables model to first explore optimal solution in current data without constrains, and subsequently perform constrained refinement around local optimal. 

Specifically, during the first stage, both regularization terms are deactivated, allowing the model to freely adapt and approach a local optima region for current task.
The loss function for Task~$i$ in this stage is defined as follow:
\begin{equation}
    \mathcal{L}_1 = \frac{1}{|\mathcal{D}_i|} \sum_{(x_k, y_k) \in \mathcal{D}_i} L_{ce}(f_{\theta_{i, 1}'}(x_k), y_k).
    \label{loss_1}
\end{equation}
Here, $\theta_{i, 1}' = W_0 + \theta_{i, 1}$, where $\theta_{i, 1}$ is initialized from $\theta_{i-1, 1}$. For the first task, $\theta_{1, 1}$ is initialized from scratch.

During the second training stage, we simultaneously activate both fine-tuning objective and regularization to ensure parameter updates not interfering with or degrading the performance of previously learned tasks. 
The loss function for Task~$i$ in this stage is defined as follows:
\begin{equation}
    \begin{aligned}
        \mathcal{L}_2 = \frac{1}{|\mathcal{D}_i|} \sum_{(x_k, y_k) \in \mathcal{D}_i} (&L_{ce}(f_{\theta_{i, 2}'}(x_k), y_k) + \lambda_1 L_{orth}(\theta_{i, 2}) \\
        &+ \lambda_2 L_{norm}(\theta_{i, 2})),
    \end{aligned}
    \label{loss_2}
\end{equation}
where $\theta_{i, 2}' = W_0 + \sum\limits_{m=1}^{i}\theta_{m, 2}$ and $\theta_{i, 2}$ is initialized by $\theta_{i, 1}$. $L_{norm}(\theta_i)$ is the L2-regularization term, and $\lambda_1,\lambda_2$ are denoted as the weight of regularization terms.

This design enforces stability across tasks by explicitly regularizing the optimization trajectory within a subspace that preserves prior knowledge. Experimental results demonstrate that the proposed two-stage finetuning paradigm effectively safeguards model's performance, achieving a favorable trade-off between retaining past knowledge and optimizing for the current task. Our complete algorithmic procedure is summarized in Algo. \ref{algo}.

%% file: tables/Main_result.tex
\renewcommand{\arraystretch}{0.9}
\begin{table*}[t]
\begin{center}
\scalebox{0.85}{
\begin{tabular}{clc|cccccc|c}
\toprule
& Method & Replay & ImageNet-R & ArXivQA & VizWiz & IconQA & CLEVR-Math & Flickr30k & {Average}\\
\hline
& Zero-shot & - & 17.20 & 52.16 & 41.07 & 18.70 & 18.63 & 46.99 & 32.46 \\
& Multi-task & - & 89.53 & 92.20 & 62.70 & 61.86 & 70.40 & 58.49 & 72.53 \\
\hline\hline
\multirow{14}{*}{\rotatebox{90}{{Avg}}}
& Sequential Finetune & - & 49.31 & 78.40 & 50.48 & 53.44 & 55.53 &  {57.95} & 57.52  \\
& Vanilla Rehearsal & \scalebox{0.90}{\yes} & 88.71 & 86.13 & 54.15 & 61.76 & 70.53 & 58.13 & 69.90 \\

& \multicolumn{9}{>{\columncolor{gray!30}}c}{\textcolor{olive!20!black}{$\triangledown$ Architecture-based}} \\
& L2P~\cite{wang2022learning} & \scalebox{0.90}{\no} & 41.52 & 82.32 & 51.98 & 52.21 & 43.16 & 52.77 & 53.99  \\
& MoELoRA~\cite{chen2024coin} & \scalebox{0.90}{\no} & 64.49 & 82.42 & 49.54 & 56.87 & 56.35 & \textbf{58.34} & 61.33  \\
& HiDe-LLaVA~\cite{guo2025hide} & \scalebox{0.90}{\no} & {85.70} & \underline{92.70} & 54.10 & \textbf{66.87} & {59.12} & 55.15 & {68.94} \\

& \multicolumn{9}{>{\columncolor{gray!30}}c}{\textcolor{olive!20!black}{$\triangledown$ Regularization-based}} \\
& LwF~\cite{li2017learning} & \scalebox{0.90}{\no} & 55.60 & 79.86 & 53.23 & 54.87 & 56.51 & 56.34 & 59.40  \\
& EWC~\cite{kirkpatrick2017overcoming} & \scalebox{0.90}{\no} & 54.23 & 80.13 & 53.14 & 55.06 & {57.52} & 55.94 & 59.34  \\
& O-LoRA~\cite{wang2023orthogonal} & \scalebox{0.90}{\no} & 75.26 & 86.73 & {55.86} & 58.47 & 57.38 & 53.52 & 64.54  \\
\cline{2-10}
& {\textbf{Octopus (ours)\dag}} & \scalebox{0.90}{\no} & \textbf{89.69} & {91.34} & \underline{61.04} & \underline{60.18} & \textbf{68.09} & \underline{57.61} & \textbf{71.33} \\
& {\textbf{Octopus (ours)}} & \scalebox{0.90}{\no} & \underline{88.41} & \textbf{93.30} & \textbf{61.65} & {59.91} & \underline{66.32} & {56.87} & \underline{71.08} \\

\hline\hline
\multirow{14}{*}{\rotatebox{90}{{Last}}}
& Sequential Finetune & - & 37.63 & 72.33 & 43.47 & 41.70 & 35.63 & {57.95} & 48.12  \\
& Vanilla Rehearsal & \scalebox{0.90}{\yes} & 88.43 & 85.36 & 51.50 & 60.10 & 67.13 & 58.13 & 68.44 \\

& \multicolumn{9}{>{\columncolor{gray!30}}c}{\textcolor{olive!20!black}{$\triangledown$ Architecture-based}} \\
& L2P~\cite{wang2022learning} & \scalebox{0.90}{\no} & 32.73 & 80.41 & 43.72 & 42.16 & 39.25 & 52.77 & 48.51  \\
& MoELoRA~\cite{chen2024coin} & \scalebox{0.90}{\no} & 49.87 & 77.63 & 43.65 & 46.40 & 36.47 & \textbf{58.34} & 52.06  \\
& HiDe-LLaVA~\cite{guo2025hide} & \scalebox{0.90}{\no} & {80.50} & \underline{89.83} & {48.78} & \textbf{62.90} & {47.97} & 55.15 & {64.19} \\

& \multicolumn{9}{>{\columncolor{gray!30}}c}{\textcolor{olive!20!black}{$\triangledown$ Regularization-based}} \\
& LwF~\cite{li2017learning} & \scalebox{0.90}{\no} & 40.27 & 75.93 & 42.76 & 44.38 & 37.43 & 56.34 & 49.52  \\
& EWC~\cite{kirkpatrick2017overcoming} & \scalebox{0.90}{\no} & 39.05 & 77.88 & 43.24 & 45.33 & 39.72 & 55.94 & 50.20  \\
& O-LoRA~\cite{wang2023orthogonal} & \scalebox{0.90}{\no} & 69.36 & 82.42 & 48.64 & 53.66 & 42.53 & 53.52 & 58.36  \\
\cline{2-10}
& {\textbf{Octopus (ours)\dag}} & \scalebox{0.90}{\no} & \textbf{88.80} & \underline{89.83} & \underline{56.44} & {59.90} & \textbf{70.60} & \underline{57.11} & \underline{70.45} \\
& {\textbf{Octopus (ours)}} & \scalebox{0.90}{\no} & \underline{88.20} & \textbf{93.03} & \textbf{58.46} & \underline{60.50} & \underline{69.00} & {56.87} & \textbf{71.01} \\

\bottomrule
\end{tabular}
}
\end{center}
\vspace{-0.5cm}
\caption{Comparison with various methods on UCIT~\cite{guo2025hide} in terms of \emph{Avg} and \emph{Last}. The best and second methods are labeled with {bold} and {underline} styles. \textit{Zero-shot} evaluates pretrained model without finetuning. \textit{Multi-task} jointly finetunes model across all datasets, whereas \textit{Sequential Finetune} adapts only one LoRA module sequentially to all tasks. These settings provide an empirical characterization of the lower bound, upper bound, and baseline for continual learning methods. \dag\ denotes the use of Historical task proxy approximation in Eq. \ref{new_orth}}
\label{tab_main}
\vspace{-0.2cm}
\end{table*}
\renewcommand{\arraystretch}{1.0}

%% file: tables/Orthogonalization.tex
\begin{table*}[t]
\begin{center}
\scalebox{0.85}{
\begin{tabular}{cccccccc|c|c}
\toprule
\multicolumn{2}{c}{Settings}  & ImageNet-R & ArXivQA & VizWiz & IconQA & CLEVR-Math & Flickr30k & \textbf{Average} & \textbf{BWT}\\
\hline
\multicolumn{10}{>{\columncolor{gray!30}}c}{\textcolor{olive!20!black}{$\triangledown$ Constrain current parameters orthogonal to \textit{what}}} \\
\multicolumn{1}{c}{\multirow{2}{*}{Prev. Param.}} & Imd. & 88.47 & 91.73 & 60.74 & 56.73 & 61.33 & 56.29 & 69.22 & \multirow{2}{*}{\textcolor{red!70!black}{\textbf{-2.51}}} \\
 & \cellcolor{gray!15}Last & \cellcolor{gray!15}86.67 & \cellcolor{gray!15}92.27 & \cellcolor{gray!15}50.62 & \cellcolor{gray!15}57.00 & \cellcolor{gray!15}57.40 & \cellcolor{gray!15}56.29 & \cellcolor{gray!15}66.71 &  \\
\hline
\multicolumn{1}{c}{\multirow{2}{*}{\textbf{GPWC}}} & Imd. & 88.47 & 93.13 & 61.04 & 60.47 & 63.63 & 56.87 & \textbf{70.60} & \multirow{2}{*}{\textcolor{green!50!black}{\textbf{+0.41}}} \\
 & \cellcolor{gray!15}Last & \cellcolor{gray!15}88.20 & \cellcolor{gray!15}93.03 & \cellcolor{gray!15}58.46 & \cellcolor{gray!15}60.50 & \cellcolor{gray!15}69.00 & \cellcolor{gray!15}56.87 & \cellcolor{gray!15}71.01 &  \\
\hline
\multicolumn{1}{c}{\multirow{2}{*}{Prev. Param. \& GPWC}} & Imd. & 88.47 & 93.20 & 60.50 & 60.77 & 63.60 & 56.99 & 70.59 & \multirow{2}{*}{\textcolor{green!50!black}{\textbf{+0.45}}} \\
 & \cellcolor{gray!15}Last & \cellcolor{gray!15}88.27 & \cellcolor{gray!15}92.90 & \cellcolor{gray!15}59.01 & \cellcolor{gray!15}60.27 & \cellcolor{gray!15}68.83 & \cellcolor{gray!15}56.99 & \cellcolor{gray!15}\textbf{71.04} &  \\

\hline
\multicolumn{10}{>{\columncolor{gray!30}}c}{\textcolor{olive!20!black}{$\triangledown$ Compute Gradients on \textit{what}}} \\
\multicolumn{1}{c}{\multirow{2}{*}{\textbf{Curr. Task (GPWC)}}} & Imd. & 88.47 & 93.13 & 61.04 & 60.47 & 63.63 & 56.87 & 70.60 & \multirow{2}{*}{\textcolor{green!50!black}{\textbf{+0.41}}} \\
 & \cellcolor{gray!15}Last & \cellcolor{gray!15}88.20 & \cellcolor{gray!15}93.03 & \cellcolor{gray!15}58.46 & \cellcolor{gray!15}60.50 & \cellcolor{gray!15}69.00 & \cellcolor{gray!15}56.87 & \cellcolor{gray!15}\textbf{71.01} &  \\
\hline
\multicolumn{1}{c}{\multirow{2}{*}{Prev. Task}} & Imd. & 88.47 & 93.53 & 61.53 & 61.67 & 65.70 & 56.31 & \textbf{71.20} & \multirow{2}{*}{\textcolor{red!70!black}{\textbf{-8.95}}} \\
 & \cellcolor{gray!15}Last & \cellcolor{gray!15}86.50 & \cellcolor{gray!15}73.43 & \cellcolor{gray!15}51.47 & \cellcolor{gray!15}58.57 & \cellcolor{gray!15}46.40 & \cellcolor{gray!15}57.15 & \cellcolor{gray!15}62.25 &  \\

\bottomrule
\end{tabular}
}
\end{center}
\vspace{-0.5cm}
\caption{Effectiveness of History-Free Gradient Orthogonalization. We compare the performance of three orthogonalization strategies (orthogonal to param., \textbf{GPWC} or both) and that of two type of gradients (\textbf{GPWC} and gradients of prev. param. on prev. task). Our method is emphasized in bold for clarity. We report the performance on each dataset under different settings in terms of \textit{Imd.} and \textit{Last}. }
\label{tab_orth}
\vspace{-0.0cm}
\end{table*}

%% file: tables/Two-stage_training.tex
\begin{table*}[t]
\begin{center}
\scalebox{0.85}{
\begin{tabular}{cccccccc|c|c}
\toprule
\multicolumn{2}{c}{Method}  & ImageNet-R & ArXivQA & VizWiz & IconQA & CLEVR-Math & Flickr30k & \textbf{Average} & \textbf{BWT}\\
\hline
\multicolumn{1}{c}{\multirow{2}{*}{\textbf{w/ two-stage finetuning}}} & Imd. & 88.47 & 93.13 & 61.04 & 60.47 & 63.63 & 56.87 & \textbf{70.60} & \multirow{2}{*}{\textcolor{green!50!black}{\textbf{+0.41}}} \\
 & \cellcolor{gray!15}Last & \cellcolor{gray!15}88.20 & \cellcolor{gray!15}93.03 & \cellcolor{gray!15}58.46 & \cellcolor{gray!15}60.50 & \cellcolor{gray!15}69.00 & \cellcolor{gray!15}56.87 & \cellcolor{gray!15}\textbf{71.01} &  \\
\hline
\multicolumn{1}{c}{\multirow{2}{*}{w/o two-stage finetuning}} & Imd. & 87.47 & 92.53 & 48.61 & 40.70 & 49.90 & 55.62 & 62.47 & \multirow{2}{*}{\textcolor{red!70!black}{\textbf{-1.29}}} \\
 & \cellcolor{gray!15}Last & \cellcolor{gray!15}86.90 & \cellcolor{gray!15}92.20 & \cellcolor{gray!15}45.96 & \cellcolor{gray!15}38.90 & \cellcolor{gray!15}47.53 & \cellcolor{gray!15}55.62 & \cellcolor{gray!15}61.18 &  \\
\bottomrule
\end{tabular}
}
\end{center}
\vspace{-0.5cm}
\caption{Effectiveness of two-stage finetuning strategy. We report the performance on each dataset under both w/ and w/o two-stage fine-tuning settings in terms of \textit{Imd.} and \textit{Last}.}
\label{tab_two_stage}
\vspace{-0.3cm}
\end{table*}

%% file: sec/4_experiments.tex
\section{Experiments}
\label{sec:experiments}
\begin{figure}
    \centering
    \includegraphics[width=\linewidth]{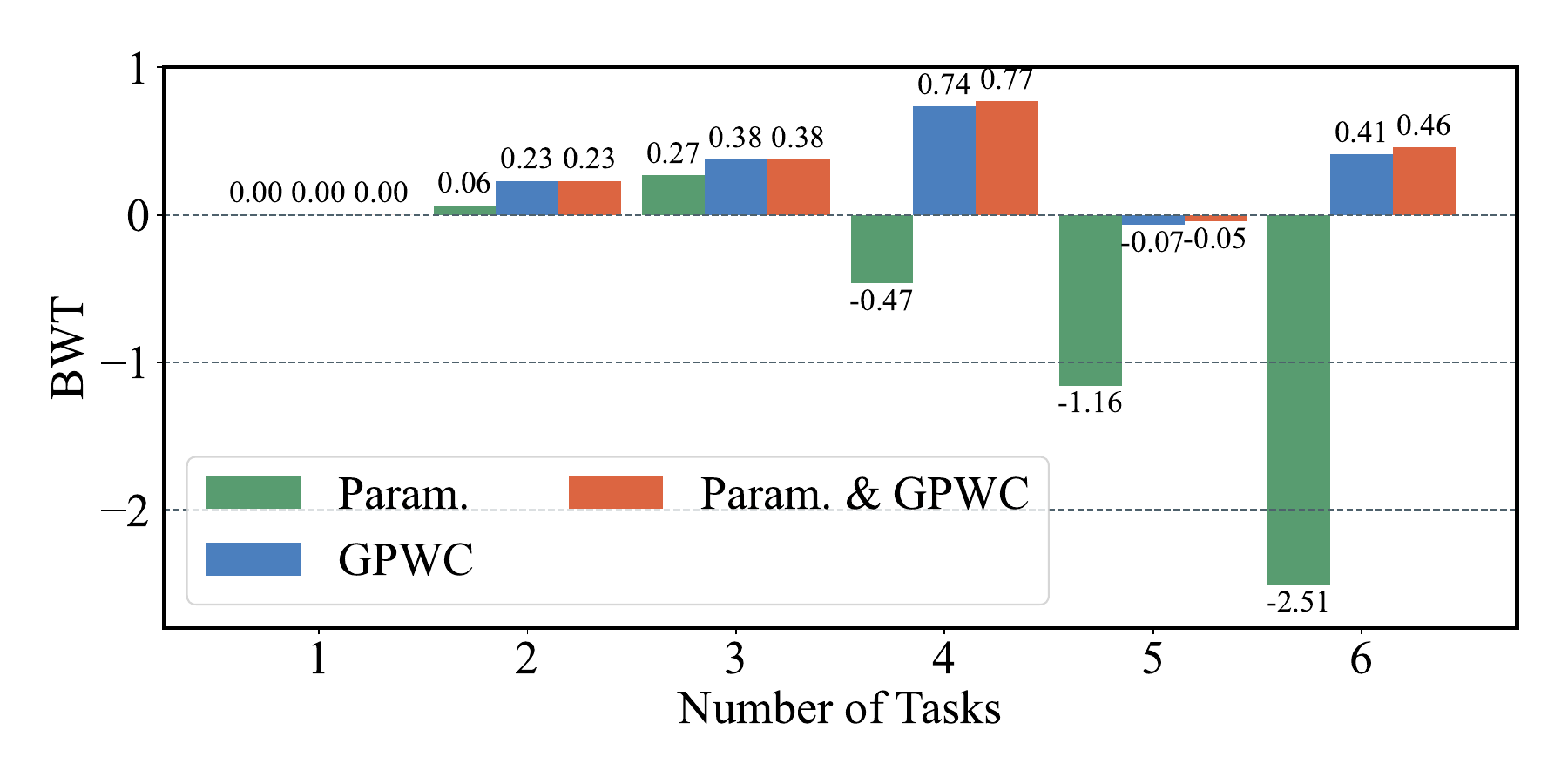}
    \vspace{-0.8cm}
    \caption{Backward transfer (BTW) of model after fine-tuning on each task with different orthogonalization method.}
    \label{fig_orth}
    \vspace{-0.3cm}
\end{figure}
\subsection{{Experimental Setup}}

\vspace{5pt}\noindent\textbf{{Benchmark.}}
To evaluate the effectiveness of \our, we conduct experiments on UCIT~\cite{guo2025hide}, which is specifically designed to assess the continual learning capability of MLLMs under realistic, instruction-driven scenarios. It consists of a sequence of multimodal instruction datasets covering diverse visual domains and linguistic tasks, including visual question answering, caption generation, and mathematical reasoning. Each task introduces novel visual and textual distributions, thereby inducing significant domain shifts and catastrophic forgetting challenges.

\vspace{5pt}\noindent\textbf{Metrics.}
Following UCIT~\cite{guo2025hide}, we adopt \textit{Last} and \textit{Avg} metrics to evaluate the continual learning performance of MLLMs. \textit{Last} denotes the average accuracy over all tasks after sequentially learning the entire task stream, while \textit{Avg}, on the other hand, represents the mean accuracy across all tasks throughout the training process. In addition, we adopt \textit{Imd.} in several experiments to denote performance of a given task immediately after fine-tuning in the sequence learning process, which to some extent reflects the upper bound of the performance of that task during continual learning. We also report Backward Transfer (BWT)~\cite{lopez2017gradient} in several experiments, which measures the average performance degradation on previous tasks and thereby reflects the degree of forgetting exhibited by continual learning method. For VQA tasks, we employ accuracy as the evaluation criterion, while for captioning tasks, we report the average score over multiple metrics, including BLEU-1-4~\cite{papineni2002bleu}, METEOR~\cite{denkowski2014meteor}, ROUGE-L~\cite{lin2004rouge}, and CIDEr~\cite{vedantam2015cider}.

\vspace{5pt}\noindent\textbf{{Implementation details.}}
Following UCIT~\cite{guo2025hide}, we use LLaVA-v1.5-7b~\cite{liu2024improved} as the base multimodal model 
and embed LoRA modules in all linear layers of language model.
In practice, we set $D_{i1} = D_{i}$ and select a small subset of $D_i$ as $D_{i2}$ to enforce constraints. The number of training epochs for all tasks is set to 1.
We set batch size to 16 for all methods and run experiments on 4 $\times$ H20/P800 GPUs.

\subsection{{Main Results}}
We conduct comprehensive evaluations of \our\ on UCIT~\cite{guo2025hide}, comparing it against a diverse set of existing methods, as summarized in Tab. \ref{tab_main} and Fig. \ref{fig_radar}. For the sake of fairness, we compare \our\ with only architecture-based and regularization-based approaches, while including the vanilla rehearsal method as a reference.

The quantitative results presented in Tab. \ref{tab_main} demonstrate our proposed \our\ achieves state-of-the-art (SOTA) performance and outperforms the best previous methods by 2.14\% and 6.82\% in \textit{Avg} and \textit{Last} metrics, respectively. Specifically, on one hand, \our\ demonstrates substantial improvements over existing regularization-based methods such as EWC~\cite{kirkpatrick2017overcoming}, LwF~\cite{li2017learning} and OLoRA~\cite{wang2023orthogonal}. \our\ achieves 6.54\% and 12.65\% performance improvement over OLoRA in terms of \textit{Avg} and \textit{Last}, which indicates that our HiFGO exhibits more effective weight disentanglement compared to orthogonalization based on weight space. 
On the other hand, \our\ significantly outperforms other MoE-based approaches, such as MoELoRA~\cite{chen2024coin} and HiDe-LLaVA~\cite{guo2025hide}, demonstrating that our method achieves superior mitigation of inter-task parameter interference while maintaining high inference efficiency. 

\captionsetup[subfigure]{skip=0pt}
\begin{figure}
\vspace{-0.2cm}
\centering
\subfloat[\label{fig_order}]{
		\includegraphics[width=0.48\linewidth]{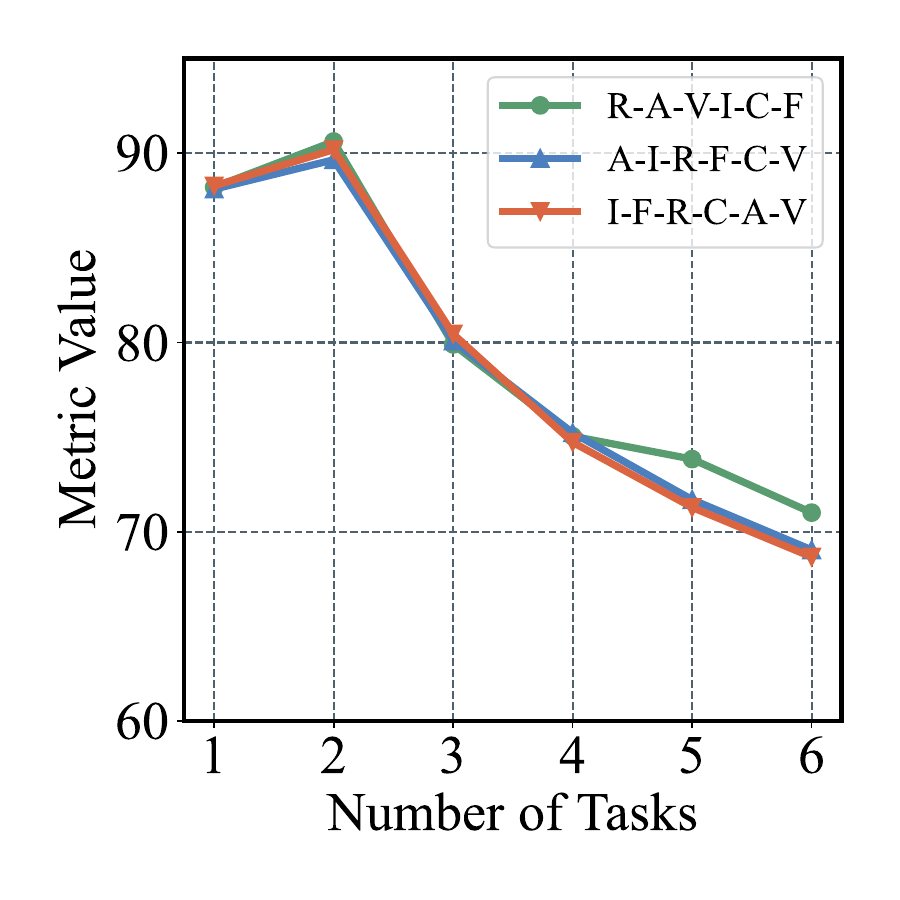}}
\hfill
\subfloat[\label{fig_two_stage}]{
		\includegraphics[width=0.48\linewidth]{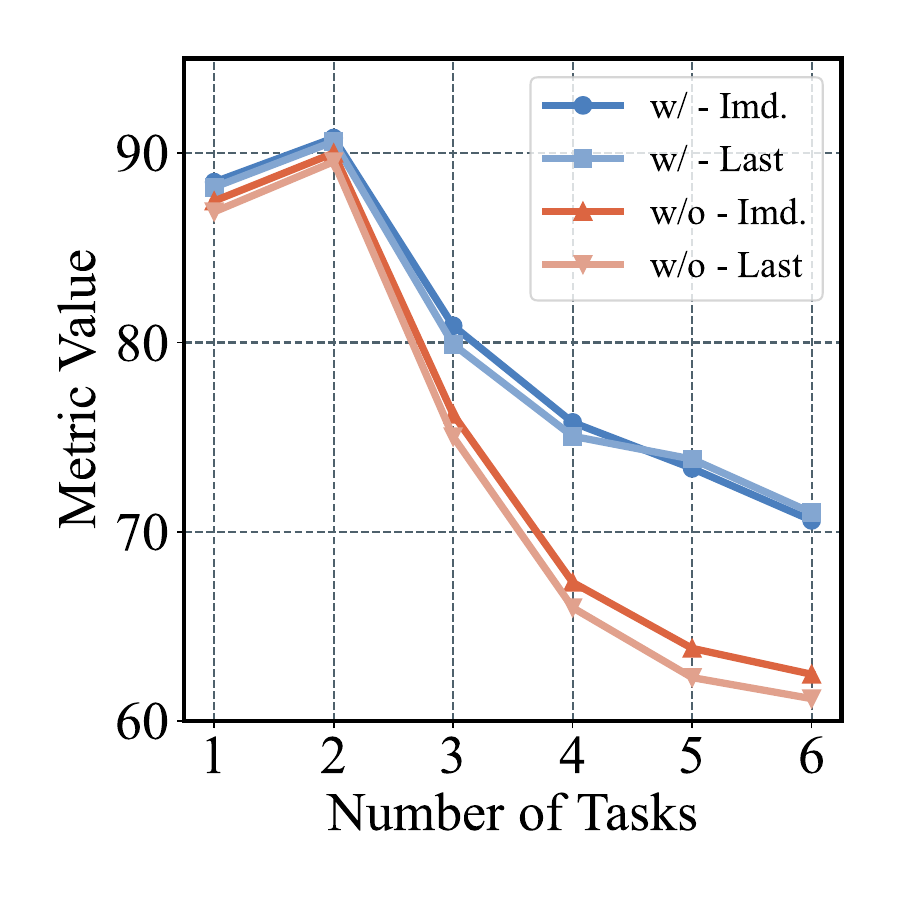}}
\vspace{-0.3cm}
\caption{Average performance after fine-tuning on each task under different orderings \textbf{(a)} or w/ and w/o two-stage finetuning \textbf{(b)}.}
\vspace{-0.6cm}
\end{figure}

Moreover, it is particularly noteworthy that \our\ even surpasses the vanilla rehearsal baseline, which explicitly constructs a replay buffer to store and reuse data samples from previous tasks. It underscores the ability of our method to achieve a more favorable plasticity–stability trade-off without relying on additional memory buffers. The explicit disentanglement of task representations also facilitates positive inter-task transfer, as evidenced by the ArXivQA \textit{Last} metric exceeding that of \textit{Multi-task}, while rehearsal-based strategies often suffer from task interference and lead to suboptimal performance. 

\subsection{Model Analysis}
\input{tables/Regularization}
\input{tables/Task-order}

\vspace{5pt}\noindent\textbf{Effectiveness of History-Free Gradient Orthogonalization.}
Tab. \ref{tab_orth} presents the comparison results obtained under different orthogonalization methods (orthogonal to previous parameters, \textbf{GPWC} or both) and settings (compute gradient of previous parameters on \textbf{current} / previous dataset), demonstrating that our proposed HiFGO exhibits a substantial performance advantage over existing parameter-space or history-based gradient-space orthogonalization.

Fig. \ref{fig_orth} illustrates the average forgetting observed during sequential learning process, quantified using BWT. The results reveal that HiFGO exhibits substantially stronger resistance to forgetting compared to parameter orthogonalization (\textcolor{green!50!black}{\textbf{+0.41}} v.s. \textcolor{red!70!black}{\textbf{-2.51}} BWT), and even yields positive backward transfer (\textcolor{green!50!black}{\textbf{+0.41}} BWT), which suggests that, the acquisition of a new task not only mitigates the degradation of performance on previous tasks but can further improve performance through the incorporation of newly acquired knowledge. within sequential learning process. Furthermore, jointly enforcing both forms of orthogonal constraints yields no additional gains, underscoring that HiFGO constitutes a more principled and inherently effective constraint mechanism compared to parameter orthogonalization.

Comparison in Tab. \ref{tab_orth} between HiFGO and history-based gradient orthogonalization reveals that the latter exhibits substantial forgetting (\textcolor{red!70!black}{\textbf{–8.95}} BWT), which may stem from the fact that previous tasks have already converged to local optima, where gradient directions become less informative and highly oscillatory. In contrast, the GPWC employed in HiFGO provides a more faithful characterization of inter-task interference, thereby playing a substantially more effective role in mitigating catastrophic forgetting.

\vspace{5pt}\noindent\textbf{{Effectiveness of the two-stage finetuning strategy.}}
Tab. \ref{tab_two_stage} reports effectiveness of our proposed two-stage finetuning strategy, indicating that our two-stage finetuning strategy yields a substantial influence on the \textit{Imd.} and \textit{Last} metrics, while exerting a relatively minor effect on BWT. It suggests that our two-stage finetuning strategy effectively bolsters model's efficacy during fine-tuning phase of individual task, elevates the performance upper bound in sequential learning, and thereby enhances overall performance. Fig. \ref{fig_two_stage} provides additional support by illustrating average \textit{Imd.} and \textit{Last} across varying task counts. Specifically, our two-stage finetuning strategy exerts a significant impact on the respective performance of \textit{Imd.} and \textit{Last}, but does not significantly affect the gap between them.

\vspace{5pt}\noindent\textbf{{Effectiveness of norm regularization.}}
As mentioned in Sec. \ref{two_stage_training}, norm regularization would effectively mitigates perturbations induced by high-order terms in Eq. \ref{eq:taylor}, yet it moderately degrades model’s finetuning efficacy. This observation is substantiated by Tab. \ref{tab_reg}: stronger norm regularization yields superior performance in alleviating catastrophic forgetting and even facilitates more pronounced positive backward transfer, but induces a marginal drop in the \textit{Imd.} metric. Consequently, a trade-off must be struck between these two objectives—an optimal balance between plasticity and stability is achieved via an appropriate norm regularization strength (we choose $\lambda = 1e-2$ in practice). 

\vspace{5pt}\noindent\textbf{{Influence of task ordering.}}
\label{ablation_on_task_order}
For clarity, we represent each dataset by a letter; for instance, I-F-R-C-A-V denotes the learning sequence IconQA $\rightarrow$ Flickr30k $\rightarrow$ ImageNet-R $\rightarrow$ CLEVR-Math $\rightarrow$ ArXivQA $\rightarrow$ VizWiz.
Tab. \ref{tab_order} presents the performance of of sequential learning under varying task orderings, in terms of \textit{Last} and \textit{Avg}, and Fig. \ref{fig_order} illustrates accumulative average post–sequential-learning performance across varying numbers of tasks, where task count $i$ corresponds to average performance over the first $i$ tasks of R-A-V-I-C-F. 
Results show that \our\ demonstrates remarkable insensitivity to task order, with performance remaining highly consistent across different permutations or number of tasks. These observations indicate that \our\ exhibits strong robustness and ensures stable, order-invariant performance throughout the continual learning process.

%% file: tables/Regularization.tex
\begin{table*}[t]
\begin{center}
\setlength{\tabcolsep}{4pt}
\scalebox{0.85}{
\begin{tabular}{ccccccccc|c|c}
\toprule
\multicolumn{3}{c}{Method}  & ImageNet-R & ArXivQA & VizWiz & IconQA & CLEVR-Math & Flickr30k & \textbf{Average} & \textbf{BWT}\\
\hline
\multicolumn{1}{c}{\multirow{2}{*}{w/o norm-regularization}}  & \multicolumn{1}{c}{\multirow{2}{*}{-}} & Imd. & 89.03 & 93.37 & 63.31 & 65.03 & 73.07 & 56.33 & \textbf{73.36} & \multirow{2}{*}{\textcolor{red!70!black}{\textbf{-8.59}}} \\
&  & \cellcolor{gray!15}Last & \cellcolor{gray!15}81.50 & \cellcolor{gray!15}81.43 & \cellcolor{gray!15}47.65 & \cellcolor{gray!15}57.57 & \cellcolor{gray!15}63.27 & \cellcolor{gray!15}57.25 & \cellcolor{gray!15}64.77 &  \\
\hline
\multicolumn{1}{c}{\multirow{8}{*}{\textbf{w/ norm-regularization}}} 
& \multicolumn{1}{c}{\multirow{2}{*}{$\lambda=2e-3$}} & Imd. & 88.73 & 92.03 & 61.90 & 62.30 & 70.30 & 56.78 & 72.01 & \multirow{2}{*}{\textcolor{red!70!black}{\textbf{-3.36}}} \\
&  & \cellcolor{gray!15}Last & \cellcolor{gray!15}85.07 & \cellcolor{gray!15}86.47 & \cellcolor{gray!15}54.93 & \cellcolor{gray!15}61.63 & \cellcolor{gray!15}67.07 & \cellcolor{gray!15}56.78 & \cellcolor{gray!15}68.65 &  \\
& \multicolumn{1}{c}{\multirow{2}{*}{$\lambda=5e-3$}} & Imd. & 88.23 & 93.87 & 62.34 & 63.37 & 68.27 & 57.13 & 72.20 & \multirow{2}{*}{\textcolor{red!70!black}{\textbf{-1.48}}} \\
&  & \cellcolor{gray!15}Last & \cellcolor{gray!15}87.40 & \cellcolor{gray!15}91.23 & \cellcolor{gray!15}58.20 & \cellcolor{gray!15}62.13 & \cellcolor{gray!15}68.23 & \cellcolor{gray!15}57.13 & \cellcolor{gray!15}70.72 &  \\
& \multicolumn{1}{c}{\multirow{2}{*}{$\lambda=1e-2$}} & Imd. & 88.47 & 93.13 & 61.04 & 60.47 & 63.63 & 56.87 & 70.60 & \multirow{2}{*}{\textcolor{green!50!black}{\textbf{+0.41}}} \\
&  & \cellcolor{gray!15}Last & \cellcolor{gray!15}88.20 & \cellcolor{gray!15}93.03 & \cellcolor{gray!15}58.46 & \cellcolor{gray!15}60.50 & \cellcolor{gray!15}69.00 & \cellcolor{gray!15}56.87 & \cellcolor{gray!15}\textbf{71.01} &  \\
& \multicolumn{1}{c}{\multirow{2}{*}{$\lambda=3e-2$}} & Imd. & 87.93 & 89.60 & 54.92 & 43.67 & 51.57 & 57.01 & 64.12 & \multirow{2}{*}{\textcolor{green!50!black}{\textbf{+2.91}}} \\
&  & \cellcolor{gray!15}Last & \cellcolor{gray!15}88.07 & \cellcolor{gray!15}92.93 & \cellcolor{gray!15}49.46 & \cellcolor{gray!15}54.57 & \cellcolor{gray!15}60.17 & \cellcolor{gray!15}57.01 & \cellcolor{gray!15}67.03 &  \\
\bottomrule
\end{tabular}
}
\end{center}
\vspace{-0.4cm}
\caption{Effectiveness of norm regularization. We report performance under both w/ and w/o settings in terms of \textit{Imd.} and \textit{Last}.}
\label{tab_reg}
\vspace{-0.2cm}
\end{table*}

%% file: tables/Task-order.tex
\begin{table}[t]
\small
	\centering
    \scalebox{0.9}{
	\begin{tabular}{c|ccc}
		\toprule	
        Order & R-A-V-I-C-F & A-I-R-F-C-V & I-F-R-C-A-V \\
		\hline
            Last & \textbf{71.01} & 69.04 & 69.16 \\
            \cellcolor{gray!15}Avg & \cellcolor{gray!15}\textbf{71.08} & \cellcolor{gray!15}70.04 & \cellcolor{gray!15}70.00 \\
		\bottomrule
	\end{tabular}
    }
\vspace{-0.15cm}
\caption{Results of different task orders on UCIT benchmark. We adopt an abbreviation scheme to simplify the representation of task sequence notation, as explained in Sec. \ref{ablation_on_task_order}}
\label{tab_order}
\vspace{-0.5cm}
\end{table}

%% file: sec/5_conclusion.tex
\section{Conclusion}
\label{sec:conclusion}
We propose \our, a two-stage continual learning framework based on history-free gradient orthogonalization. Specifically, our proposed HiFGO effectively mitigates catastrophic forgetting without respect to historical data, while our two-stage finetuning strategy achieves an effective trade-off between plasticity and stability. Experiments on UCIT corroborate that \our\ attains state-of-the-art (SOTA) performance, outperforming the prior SOTA by 2.14\% and 6.82\% in terms of the \textit{Avg} and \textit{Last}, respectively. 
Due to space constraints, our limitations: upper bound on the number of tasks and performance degradation on highly analogous tasks are detailed in Supplementary Material.

%% file: sec/6_Acknowledgement.tex
\paragraph{Acknowledgments.}
This work was supported in part by NSFC (62322113, 62376156), Shanghai Municipal Science and Technology Major Project (2025SHZDZX025G15, 2021SHZDZX0102), and the Fundamental Research Funds for the Central Universities.
We thank Kunlunxin for their technical support in training and evaluation on P800.

%% file: sec/X_suppl.tex
\renewcommand\thesection{\Alph{section}} 
\clearpage
\setcounter{page}{1}
\maketitlesupplementary

This material provides supplementary information on the proposed \our\ framework. It presents more details on our experimental settings (Sec. \ref{Further Experimental Details}), further experimental results (Sec. \ref{Further Experimental Results}), and more analysis, visualizations, and discussions (Sec. \ref{Further Analysis}) to complement the main manuscript. Finally, we discuss the limitations of \our.

\section{Further Experimental Details}
\label{Further Experimental Details}
\subsection{Benchmarks}
In the main manuscript, we evaluate our method on UCIT~\cite{guo2025hide}, while in the supplementary material we further report its performance on CoIN~\cite{chen2024coin}. Below, we provide detailed descriptions of both evaluations.

\paragraph{UCIT~\cite{guo2025hide}}
UCIT~\cite{guo2025hide} is proposed to rigorously evaluate multimodal large language models in settings where instruction-following abilities must be incrementally acquired across heterogeneous visual–linguistic domains. This benchmark is constructed by unifying six widely used multimodal instruction datasets, including ImageNet-R~\cite{hendrycks2021imagenetr}, ArXivQA~\cite{li2024arxivqa}, VizWiz-Caption~\cite{gurari2018vizwiz}, IconQA~\cite{lu2021iconqa}, CLEVR-Math~\cite{lindstrom2022clevr} and Flickr30k~\cite{plummer2015flickr30k}, and reorganizing them into a sequential task stream. These datasets collectively cover natural-image captioning, open-ended visual question answering and mathematical reasoning. 
\paragraph{CoIN~\cite{chen2024coin}}
CoIN~\cite{chen2024coin} is introduced as a rigorous benchmark for evaluating multimodal large language models (MLLMs) under a sequential instruction‐tuning paradigm. By integrating eight distinct task categories covering multiple datasets including VQAv2~\cite{goyal2017vqav2}, VizWiz~\cite{gurari2018vizwiz}, ScienceQA~\cite{lu2022sciqa}, TextVQA~\cite{singh2019textvqa}, GQA~\cite{hudson2019gqa}, OCR-VQA~\cite{mishra2019ocr}, ImageNet~\cite{deng2009imagenet} and RECCOCO~\cite{kazemzadeh2014refcoco1, mao2016refcoco2}, CoIN~\cite{chen2024coin} captures a wide spectrum of multimodal challenges, from visual question answering and grounding to image classification and OCR-based VQA, thereby exposing models to large shifts in both visual domain and task semantics.

\input{tables/Supp_CoIN}

\captionsetup[subfigure]{skip=0pt}
\begin{figure*}
\vspace{-0.2cm}
\centering
\subfloat[\label{two_stage_1}]{
		\includegraphics[width=0.3\textwidth]{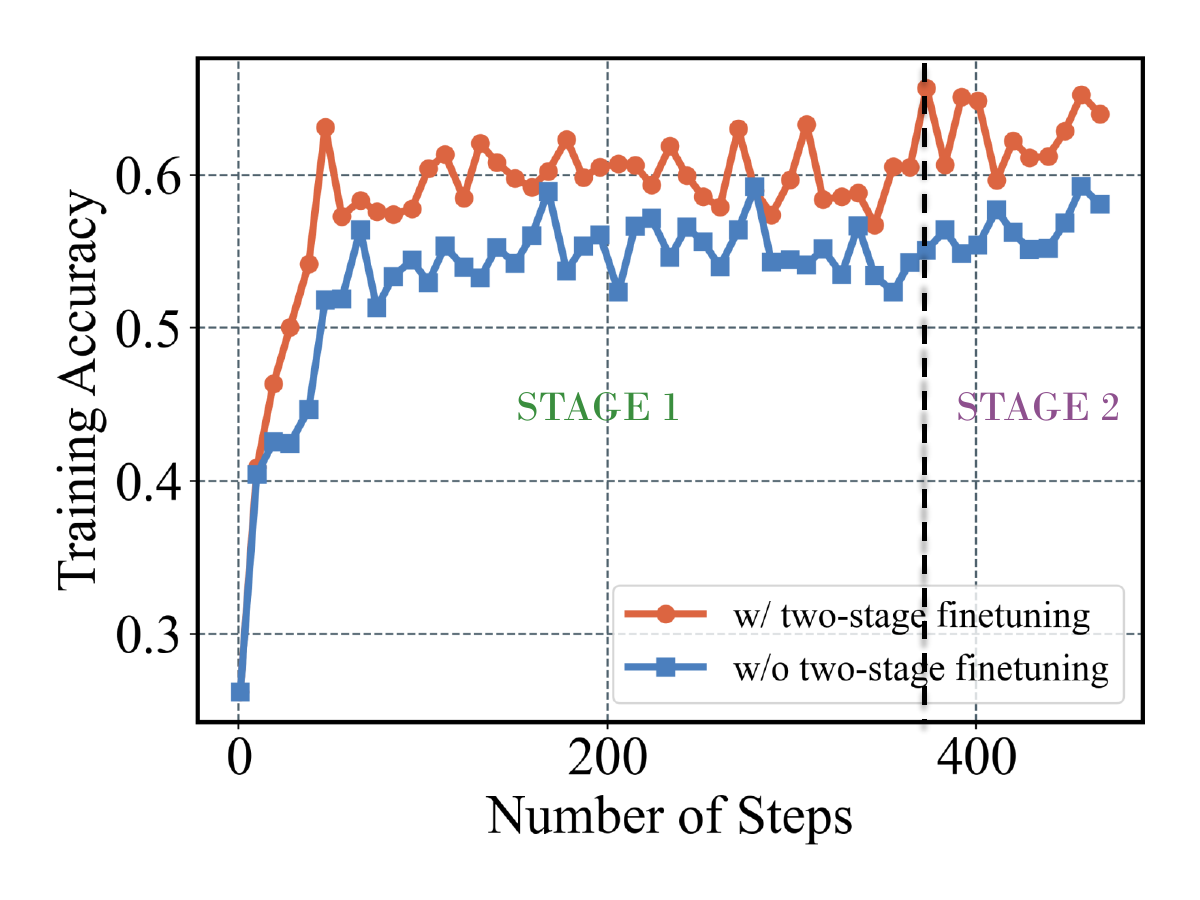}}
\hfill
\subfloat[\label{two_stage_2}]{
		\includegraphics[width=0.3\textwidth]{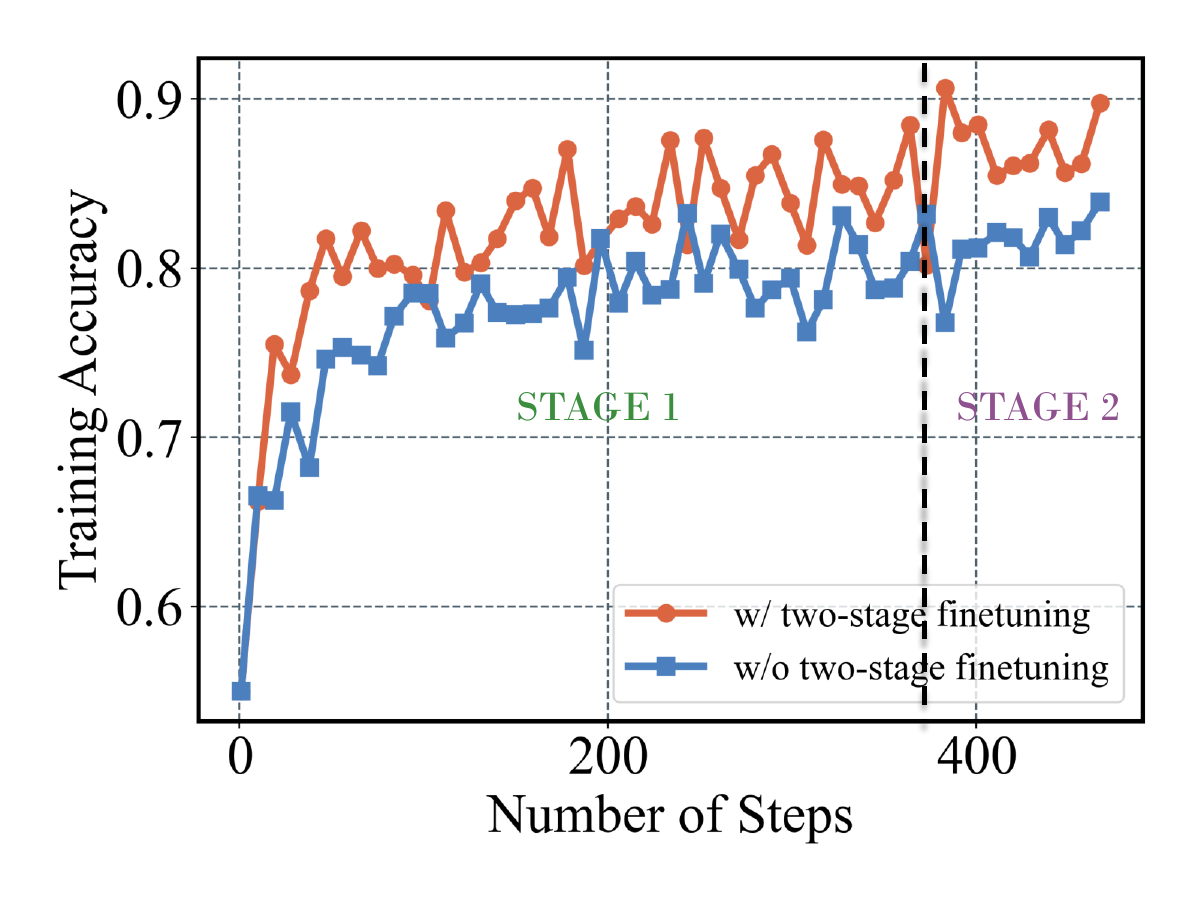}}
\hfill
\subfloat[\label{two_stage_3}]{
		\includegraphics[width=0.3\textwidth]{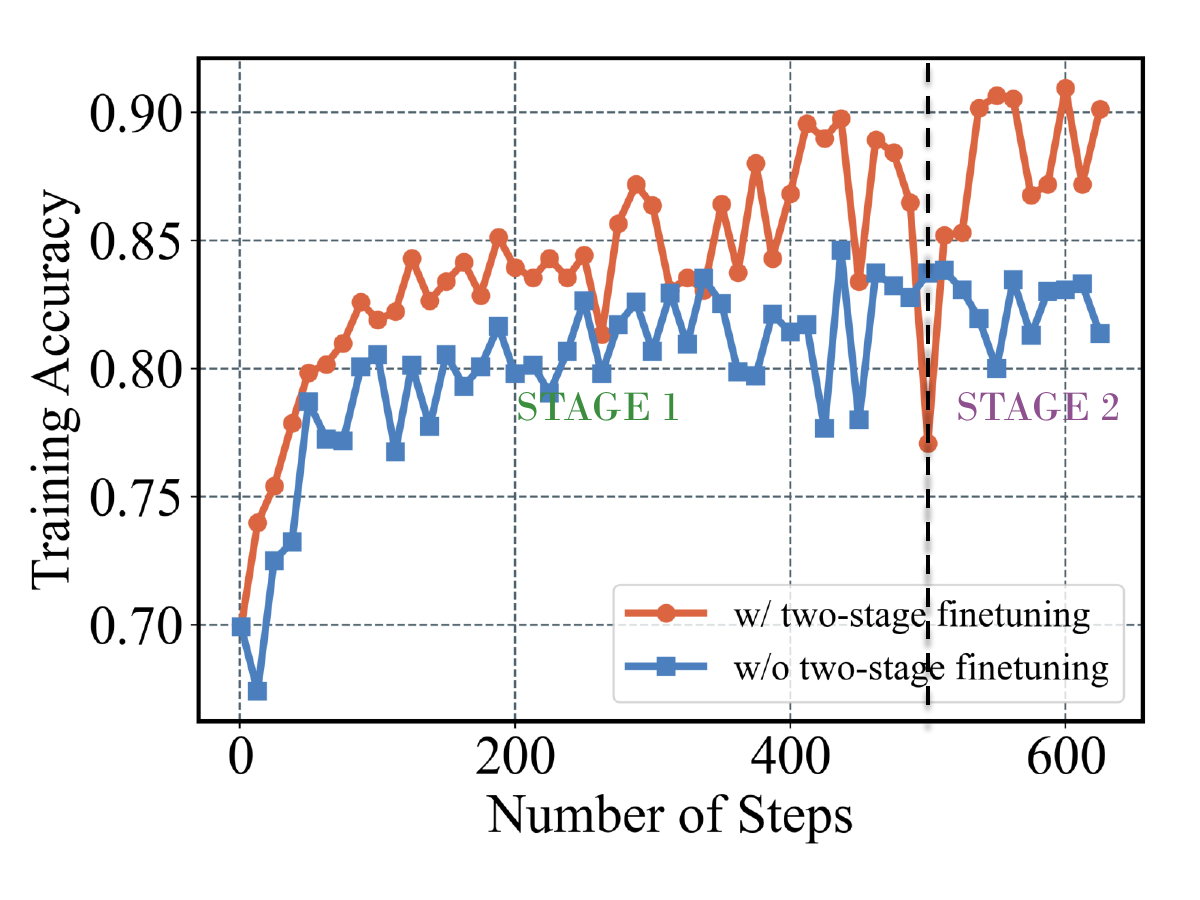}}
\vspace{-0.3cm}
\caption{Finetuning curve comparison between \textit{w/ two-stage finetuning} and \textit{w/o two-stage finetuning} on \textbf{VizWiz-Caption (a)}, \textbf{IconQA (b)} and \textbf{CLEVR-Math (c)}.}
\label{fig_two_stage_dynamic}
\vspace{-0.6cm}
\end{figure*}

\subsection{Metrics}
For standard VQA tasks, we adopt accuracy as the evaluation metric. For image captioning, we follow the protocol of UCIT and use the average score across multiple metrics, including BLEU-1-4~\cite{papineni2002bleu}, METEOR~\cite{denkowski2014meteor}, ROUGE-L~\cite{lin2004rouge}, CIDEr~\cite{vedantam2015cider} and SPICE~\cite{anderson2016spice}, as our evaluation criterion.
\paragraph{BLEU} BLEU was originally introduced in \cite{papineni2002bleu} as a metric for measuring the n-gram similarity between a predicted sentence and its reference. It provides multiple variants depending on the n-gram order, with BLEU-1, BLEU-2, BLEU-3, and BLEU-4 being the most commonly used. The computation is formulated as follows:
\begin{equation}
    p_n = \frac{\sum\limits_{C\in \{Candidates\}}\sum\limits_{n\mbox{-}gram\in C}{Count_{clip}(n\mbox{-}gram)}}{\sum\limits_{C'\in \{Candidates\}}\sum\limits_{n\mbox{-}gram'\in C'}{Count(n\mbox{-}gram')}}.
\end{equation}
\paragraph{METEOR} METEOR was proposed in \cite{denkowski2014meteor} as a text evaluation metric that integrates lemmatization, synonym matching, and a weighted precision–recall formulation to more finely assess the semantic correspondence between generated outputs and reference texts. The computation is formulated as follows:
\begin{equation}
    Score = Fmean\cdot (1 - Penalty),
\end{equation}
where:
\begin{equation}
    Fmean = \frac{10PR}{R + 9P},
\end{equation}
\begin{equation}
    Penalty = 0.5\cdot \left(\frac{\#chunks}{\#unigrams\_matched}\right)^3.
\end{equation}
\paragraph{ROUGE-L} ROUGE-L was proposed in \cite{lin2004rouge} as an automatic text evaluation metric based on longest-common-subsequence matching, which quantifies the structural overlap between a generated sequence and a reference sequence to assess their content similarity.
\paragraph{CIDEr} CIDEr was proposed in \cite{vedantam2015cider} as a TF–IDF–weighted consensus-based evaluation metric that measures the content consistency and informational relevance between generated descriptions and reference descriptions by emphasizing semantically distinctive n-grams. The computation is formulated as follows:
\begin{equation}
    CIDER(c_i, S_i) = \sum\limits_{n=1}^Nw_nCIDER_n(c_i, S_i),
\end{equation}
where:
\begin{equation}
    CIDER_N(c_i, S_i) = \frac{1}{m}\sum\limits_j{\frac{g^n(c_i)\cdot g^n(s_{ij})}{||g^n(c_i)||\cdot ||g^n(s_{ij})||}}
\end{equation}
\paragraph{SPICE} SPICE was proposed in \cite{anderson2016spice} as a scene-graph–based semantic evaluation metric that assesses the degree of semantic alignment between generated and reference descriptions by comparing their structured representations of objects, attributes, and relations. The computation is formulated as follows:
\begin{equation}
    SPICE(c, S) = F_1(c, S) = \frac{2P(c, S)R(c, S)}{P(c, S) + R(c, S)}
\end{equation}

\section{Further Experimental Results}
\label{Further Experimental Results}
We conduct extensive comparisons against a variety of methods on the CoIN~\cite{chen2024coin} benchmark, and the results are presented in Tab. \ref{tab_coin}. Compared with the current state-of-the-art (SOTA), our approach achieves improvements of 1.99\% and 2.29\% on \textit{Avg} and \textit{Last}, respectively. Consistent with the main manuscript, the competing methods are grouped into two categories: regularization-based and architecture-based.

On one hand, our method significantly outperforms other regularization-based approaches, surpassing OLoRA~\cite{wang2023orthogonal} by 4.09\% and 4.33\% on \textit{Avg} and \textit{Last}, respectively. This further demonstrates that HiFGO exhibits more effective weight disentanglement compared to orthogonalization in the weight space, and that it is applicable to diverse data and task formats with strong robustness.

On the other hand, our method also outperforms architecture-based approaches, indicating that it successfully mitigates parameter interference across tasks. In contrast, MoE tends to suffer from inaccurate task-origin prediction when processing test inputs, which limits its performance.

\section{Further Analysis}
\label{Further Analysis}
\subsection{Theoretical Analysis of GPWC}
\label{theoretical_analysis}
For $Task\ A$ and $Task\ B$ in the task sequence, we denote $\theta_A^*$ as the parameters after fine-tuning on $Task\ A$, and $\mathcal{l}(x;\theta_A^*)$ as the loss function under input $x$ and parameter $\theta_A^*$. Then, GPWC can be expressed as:
\begin{equation}
    g_{gpwc} = \mathbb{E}_{x\sim D_B}[\nabla_{\theta}\mathcal{l}(x;\theta_A^*)],
\end{equation}
where $D_B$ is the data distribution of $Task\ B$.

Let 
\begin{equation}
    \mathcal{L}_A(\theta) = \mathbb{E}_{x\in D_A}[\nabla_{\theta}\mathcal{l}(x;\theta)],
\end{equation}
\begin{equation}
    \mathcal{L}_B(\theta) = \mathbb{E}_{x\in D_B}[\nabla_{\theta}\mathcal{l}(x;\theta)],
\end{equation}
we define a family of tasks: 
\begin{equation}
    \mathcal{L}_\lambda(\theta) = \mathcal{L}_A(\theta) + \lambda(\mathcal{L}_B(\theta) - \mathcal{L}_A(\theta)),
\end{equation}
where $\lambda = 0/\lambda = 1$ represents $Task\ A/B$.

Under $\mathcal{L}_\lambda(\theta)$, the optimal parameters $\theta^*(\lambda)$ satisfy:
\begin{equation}
    \left.\frac{\partial}{\partial\theta}\mathcal{L}_\lambda(\theta)\right|_{\theta=\theta^*(\lambda)} = 0,
\end{equation}
which further implies
\begin{equation}
    \nabla_\theta\mathcal{L}_A(\theta^*(\lambda)) + \lambda(\nabla_\theta\mathcal{L}_B(\theta^*(\lambda)) - \nabla_\theta\mathcal{L}_A(\theta^*(\lambda))) = 0.
\end{equation}
Differentiating the above equation with respect to $\lambda$, we obtain
\begin{equation}
\begin{aligned}
    &\frac{\partial}{\partial\lambda}\nabla_\theta\mathcal{L}_A(\theta^*(\lambda)) + \nabla_\theta\mathcal{L}_B(\theta^*(\lambda)) - \nabla_\theta\mathcal{L}_A(\theta^*(\lambda)) \\
    + &\lambda\frac{\partial}{\partial\lambda}(\nabla_\theta\mathcal{L}_B(\theta^*(\lambda)) - \nabla_\theta\mathcal{L}_A(\theta^*(\lambda))) = 0.
\end{aligned}
\end{equation}
By substituting $\lambda=0$ (at which point $\left.\theta^*(\lambda)\right|_{\lambda=0} = \theta_A^*$ and $\nabla_\theta\mathcal{L}_A(\theta^*(\lambda))|_{\lambda=0} = \nabla_\theta\mathcal{L}_A(\theta_A^*) = 0$), we obtain
\begin{equation}
    \frac{\partial\theta^*}{\partial\lambda}\frac{\partial}{\partial\theta^*} \mathbb{E}_{x\in D_A}[\nabla_{\theta}\mathcal{l}(x;\theta_A^*)] + \mathbb{E}_{x\in D_B}[\nabla_{\theta}\mathcal{l}(x;\theta_A^*)] = 0,
\end{equation}
which implies
    $g_{GPWC} = H_A v$,
where $H_A$ denotes the Hessian matrix of $Task\ A$ evaluated at $\theta_A^*$, and
    $v=-\left.\frac{\partial\theta^*(\lambda)}{\partial\lambda}\right|_{\theta^*(\lambda) = \theta_A^*}$
corresponds to the tangent direction in parameter space induced by the data manifold of $Task\ B$, which can be interpreted as the adaptation direction of $\theta_A^*$ toward $Task\ B$.

Let $\{u_i\}$ denote the eigenvectors of $H_A$, and $v$ can be decomposed as $v = \sum\limits_{i}\gamma_iu_i$, and
\vspace{-0.2cm}
\begin{equation}
    g_{GPWC} =  H_Av = \sum\limits_{i}\lambda_i\gamma_iu_i,
\end{equation}
which is a weighted combination with the eigenvalues $\{\lambda_i\}$ of $H_A$ as weights. This indicates that GPWC primarily captures the projection of $v$ onto the high-curvature directions of $H_A$. These directions correspond to those along which updating $\theta_A^*$ under the data distribution of $Task\ B$ is more likely to degrade the performance on $Task\ A$. Therefore, constraining the parameter update direction to be orthogonal to GPWC can effectively alleviate catastrophic forgetting.

\begin{figure}
    \centering
    \includegraphics[width=\linewidth]{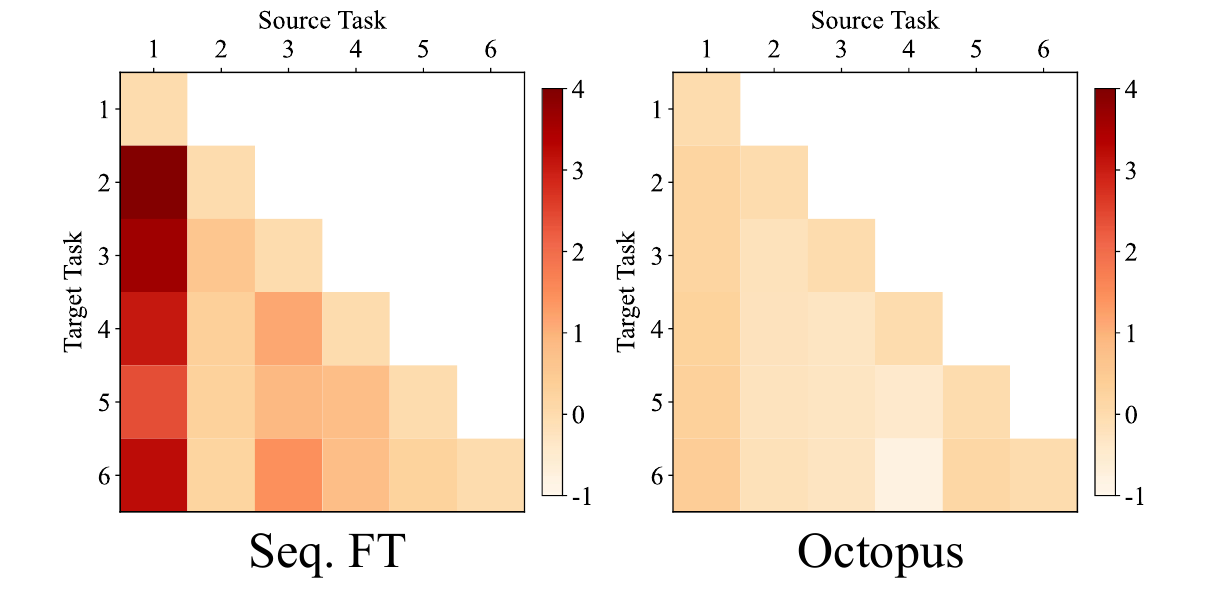}
    \caption{$\Delta\theta_{s\rightarrow t}^TH_S\Delta\theta_{s\rightarrow t}$ Comparison for Seq. FT and \our}
    \label{figure-pair}
\end{figure}

Further, we provide an empirical analysis in Fig.~\ref{figure-pair}: we compare the quadratic $\Delta\theta_{s\rightarrow t}^TH_S\Delta\theta_{s\rightarrow t}$ for Seq. FT and Octopus, where $\Delta\theta_{s\rightarrow t}$ represent parameter increment of ``source $\rightarrow$ target'', and $H_S$ is Hessian of source task.
GPWC gains a lower quadratic value, proving it can significantly mitigate the performance impact on prior tasks.

\begin{figure}
    \centering
    \includegraphics[width=0.5\linewidth]{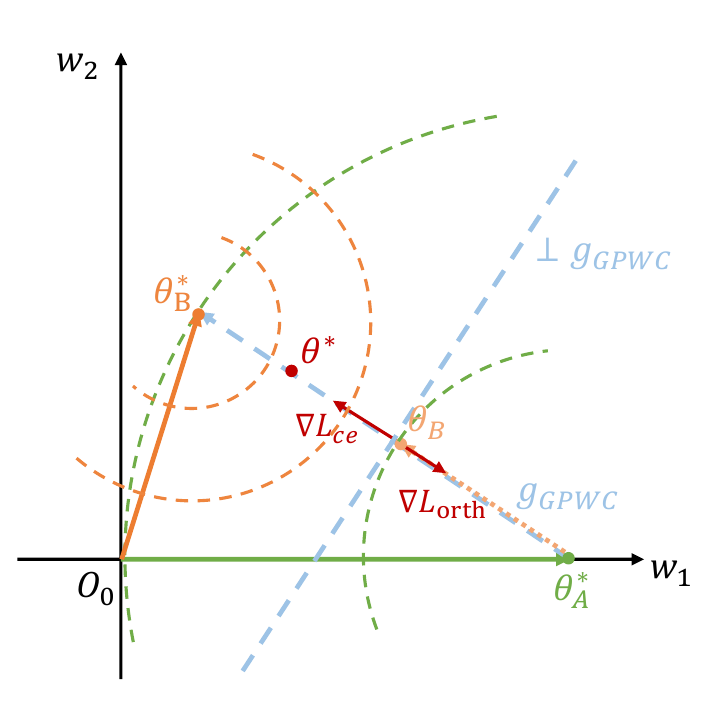}
    \caption{A toy example for GPWC.}
    \label{fig-toy_example}
\end{figure}

\paragraph{A toy example}
We provide a toy example to intuitively illustrate the role of GPWC. 
We illustrate the intuition using a simple linear regression model $y = w_1x_1 + w_2x_2$. 
We show the optimal parameters $\theta_A^*$ and $\theta_B^*$ of two linear regression tasks $T_A$ and $T_B$ in Fig. \ref{fig-toy_example}, and the blue and orange dashed curves represent the loss contours of $T_A$ and $T_B$, respectively. After the model is trained on $T_A$, the parameters converge to $\theta_A^*$. If the model is subsequently trained on $T_B$ without any constraint, the parameter update will proceed approximately along the line connecting $\theta_A^*$ and $\theta_B^*$ due to linearity, eventually converging to $\theta_B^*$. As a result, the performance on $T_A$ deteriorates significantly. Now consider the case where the GPWC constraint is introduced. When the model parameters move toward a point $\theta_B$, the GPWC direction is defined as the direction from $\theta_A^*$ to $\theta_B^*$. Notably, this direction also corresponds to the direction along which the loss on $T_A$ increases during optimization for $T_B$. Therefore, the gradient direction of $L_{orth}$ becomes opposite to that of the gradient of the task loss $L_{ce}$. Consequently, when a balance between $L_{ce}$ and $L_{orth}$ is achieved during training, the resulting parameter $\theta^*$ attains strong performance on $T_B$ while effectively preserving the performance on $T_A$.

\subsection{Example Analysis}

\begin{figure}
    \centering
    \includegraphics[width=\linewidth]{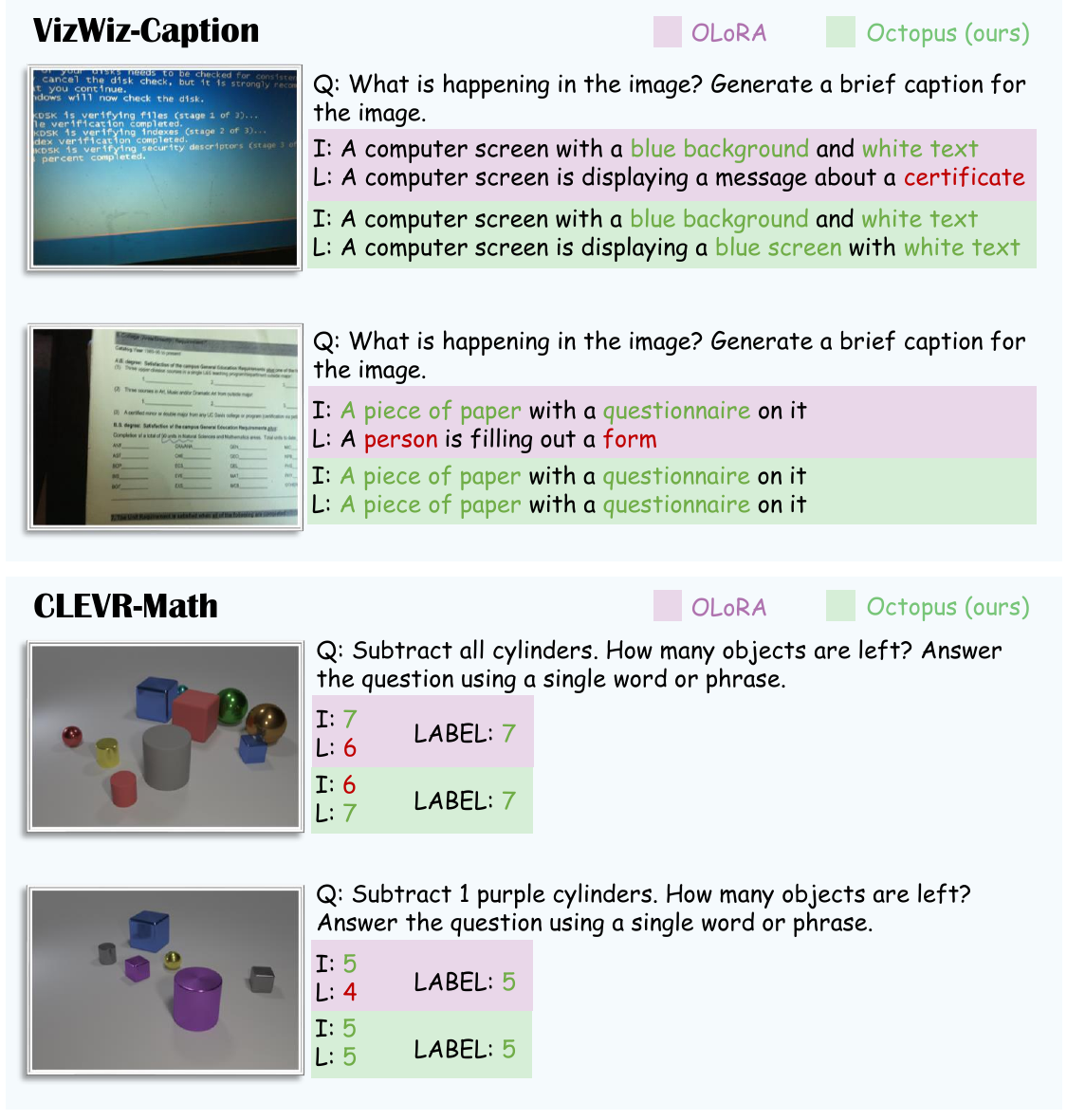}
    \vspace{-0.5cm}
    \caption{Instance-wise comparison between Octopus (ours) and OLoRA on the UCIT benchmark. Here, \textit{I} denotes \textit{Imd.} and \textit{L} denotes \textit{Last}. We illustrates the output comparison of the two methods after fine-tuning on a specific dataset and upon the completion of all training procedures.}
    \vspace{-0.5cm}
    \label{fig_visual}
\end{figure}

We provide a comparative analysis in Fig. \ref{fig_visual} of our \our\ against OLoRA~\cite{wang2023orthogonal}. \our\ demonstrates markedly superior retention of previously acquired task capabilities after sequential learning. In the image captioning task, \our\ preserves salient visual details more faithfully, while in multimodal mathematical reasoning tasks, \our\ not only reproduces the correct answers obtained at finetuning time but, in some cases, autonomously corrects errors.

\subsection{Comparison of FineTuning Dynamics for Two-Stage Strategy}
To further investigate the effectiveness of our two-stage finetuning strategy, we visualize the finetuning dynamics on several datasets in Fig. \ref{fig_two_stage_dynamic}, comparing the cases with and without the proposed strategy. As shown in the figure, during the first stage of unconstrained finetuning, the two-stage strategy converges faster and achieves a better final convergence result. In the second stage of constrained finetuning, the introduction of constraints does not substantially compromise the performance gains achieved during the first stage; after a certain number of steps, the model is still able to reach the performance level of the unconstrained setting.

In contrast, directly applying constrained finetuning from the beginning leads to a noticeable degradation in finetuning performance, limiting the model’s plasticity. Our two-stage finetuning strategy strikes an effective balance between plasticity and stability, retaining adaptation capacity while avoiding excessive performance loss.

\begin{figure}
    \centering
    \includegraphics[width=\linewidth]{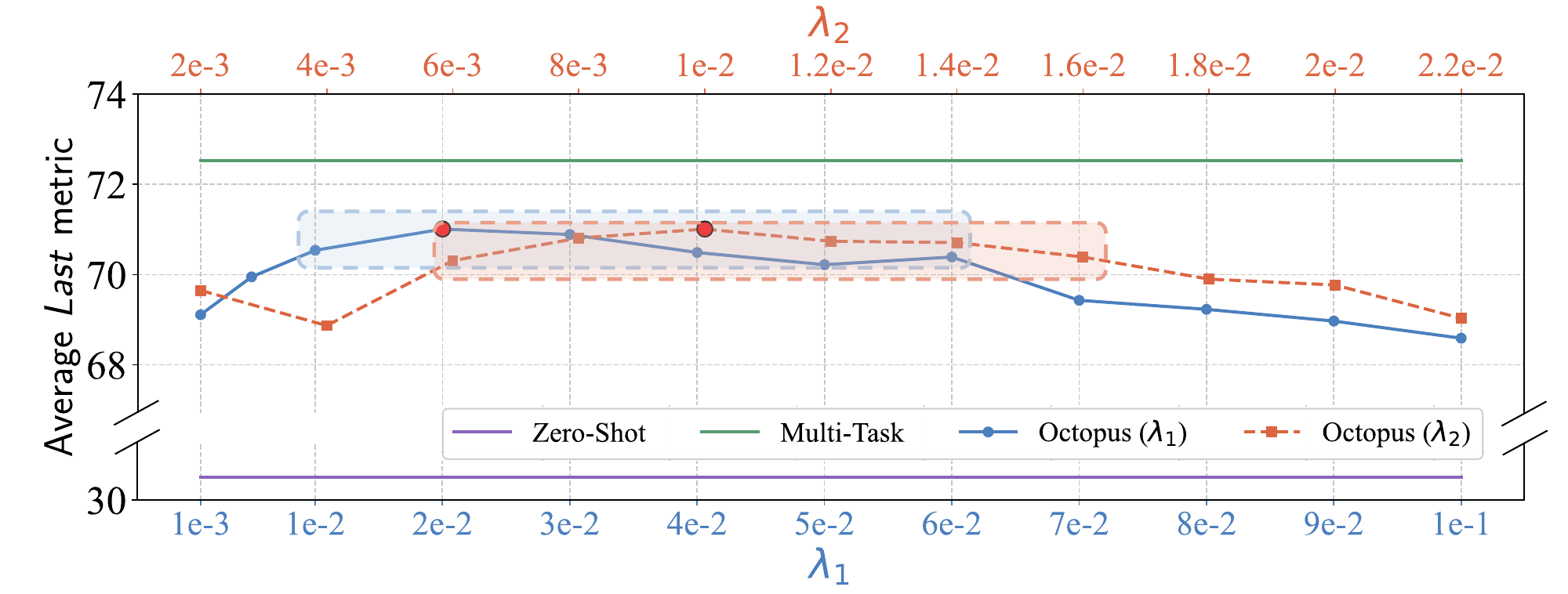}
    \caption{Sensitivity analysis of $\lambda_1$ and $\lambda_2$}
    \label{Fig_lambda}
\end{figure}

\subsection{Sensitive analysis of $\lambda_1$ and $\lambda_2$.}
\label{ablation on sensitive}
$\lambda_1$ and $\lambda_2$ are set to balance the loss magnitudes. In the experiments, we default to setting $\lambda_1 = 2 \times 10^{-2}$ and $\lambda_2 = 1 \times 10^{-2}$, and a comprehensive sensitivity analysis of the \textit{Last} metric on the UCIT dataset with respect to $\lambda_1$ and $\lambda_2$ is presented in Fig. \ref{Fig_lambda}. Results shows that the the performance remains consistently high and stable when $\lambda_1 \in [1\times 10^{-2}, 6\times 10^{-2}]$ and $\lambda_2 \in [6\times 10^{-3}, 1.6\times 10^{-2}]$. The wide effective ranges demonstrate that our method is not overly sensitive to these hyperparameters and exhibits strong robustness against their variations, which validates substantial practical applicability of \our\ .

\subsection{Extra Number of Parameters for Inference}

\begin{figure}
    \centering
    \includegraphics[width=0.95\linewidth]{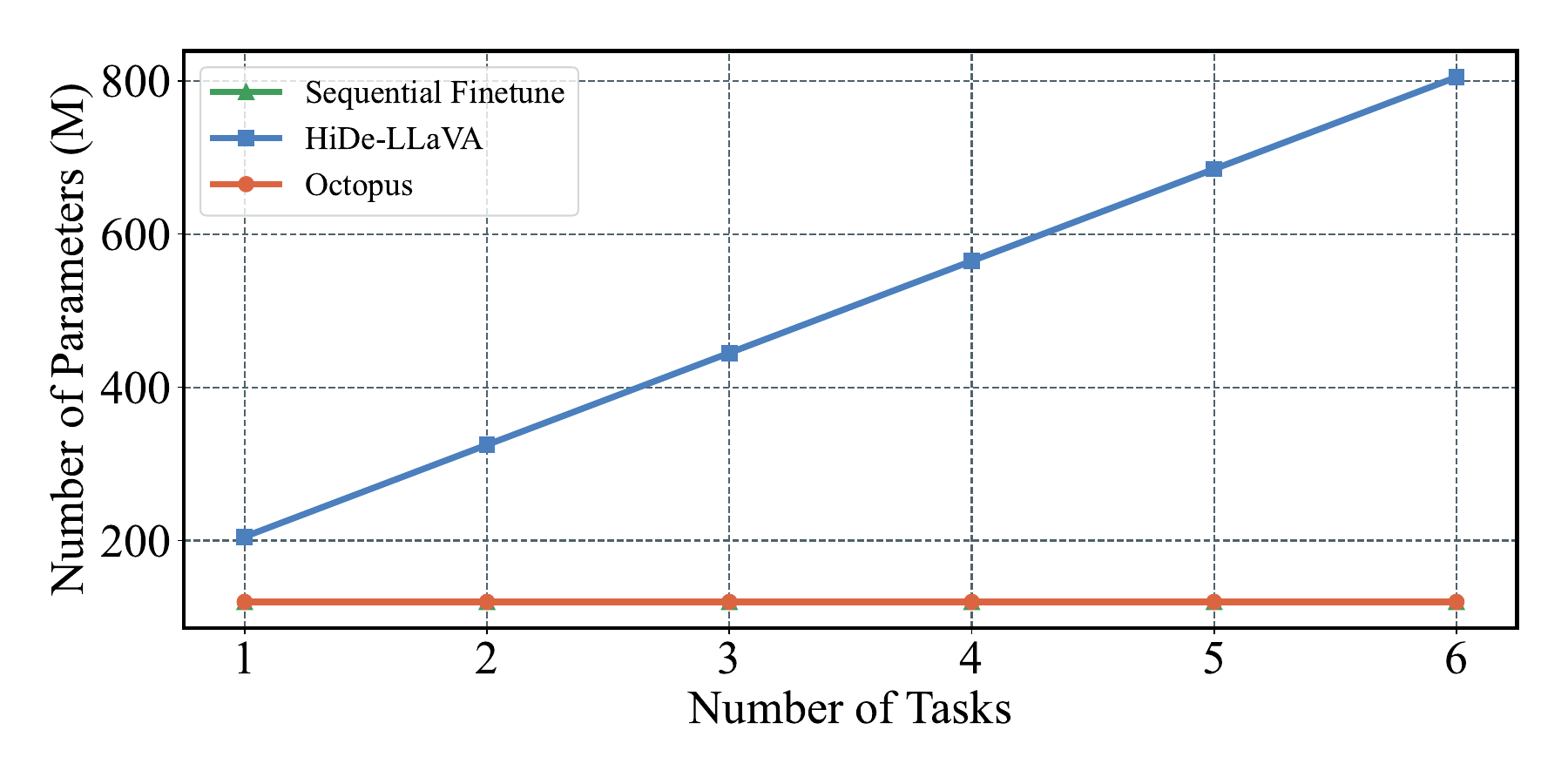}
    \caption{Comparison of the additional number of parameters required during inference.}
    \label{fig_param}
\end{figure}
In Fig. \ref{fig_param}, we compare the additional parameters introduced during inference between \our\ and the current SOTA method HiDe-LLaVA~\cite{guo2025hide} (We apply LoRA to all linear layers with rank = 48 and alpha = 96 for each method). MoE-based approaches, such as HiDe-LLaVA~\cite{guo2025hide}, require the loading of multiple expert modules—\ie, multiple LoRAs—during inference, and further necessitate an additional network component for task-ID assignment. In contrast, \our\ introduces only a single LoRA during inference, identical to standard sequential fine-tuning with LoRA.

As shown in Fig. \ref{fig_param}, compared with HiDe-LLaVA~\cite{guo2025hide}, \our\ incurs only a negligible number of additional parameters during inference, and more importantly, this overhead does not scale with the number of tasks. This design substantially reduces storage costs and improves inference efficiency.

\paragraph{Limitations}
Despite substantial improvements achieved, several limitations remain that warrant further investigation. First, similar to other LoRA-based regularization methods, our framework is constrained by the number of tasks; second, our framework would impair the performance of individual tasks for those that are highly similar in domain but differ in problem formulation. These limitations highlight the need for more sophisticated designs capable of overcoming task capacity constraints while effectively handling similar tasks. We hope our work provides valuable insights for future continual learning research for MLLMs.

%% file: tables/Supp_CoIN.tex
\renewcommand{\arraystretch}{0.9}
\begin{table*}[t]
\begin{center}
\scalebox{0.90}{
\begin{tabular}{clc|cccccccc|c}
\toprule
& Method & Replay & SciQA & Image & Viz & REC & Text & GQA & VQA & OCR & \textbf{Average}\\
\hline
& Zero-shot & - & 69.79 & 9.93 & 45.50 & 58.47 & 57.75 & 60.77 & 66.50 & 64.93 & 54.21 \\
& Multi-task & - & 82.36 & 89.63 & 52.51 & 65.83 & 61.27 & 59.93 & 65.67 & 62.03 & 67.40 \\
\hline\hline
\multirow{12}{*}{\rotatebox{90}{{Avg}}}
& Sequential Finetune & - &  64.22 & 40.13 & 43.87 & 38.32 & 55.04 & 55.89 & 60.61 & 64.78 & 52.86 \\

& \multicolumn{11}{>{\columncolor{gray!30}}c}{\textcolor{olive!20!black}{$\triangledown$ Architecture-based}} \\
& L2P~\cite{wang2022learning} & \scalebox{0.90}{\no} & 70.52 & 26.89 & 45.53 & 45.21 & 56.84 & 59.03 & 63.52 & 64.11 & 53.96 \\
& MoELoRA~\cite{chen2024coin} & \scalebox{0.90}{\no} & 68.38 & 48.50 & 44.22 & 40.23 & 55.62 & 57.04 & 62.14 & \textbf{65.75} & 55.24 \\
& HiDe-LLaVA~\cite{guo2025hide} & \scalebox{0.90}{\no} & \underline{74.92} & \underline{76.72} & \underline{51.24} & \underline{61.84} & 57.13 & \textbf{62.83} & \textbf{68.15} & 64.76 & \underline{64.70} \\

& \multicolumn{11}{>{\columncolor{gray!30}}c}{\textcolor{olive!20!black}{$\triangledown$ Regularization-based}} \\
& LwF~\cite{li2017learning} & \scalebox{0.90}{\no} & 65.20 & 40.63 & 43.22 & 40.05 & 56.23 & 54.67 & 60.64 & \underline{65.12} & 53.22  \\
& EWC~\cite{kirkpatrick2017overcoming} & \scalebox{0.90}{\no} & 65.11 & 40.89 & 44.09 & 39.67 &  54.92 & 56.03 & 61.12 & 64.55 & 53.30  \\
& O-LoRA~\cite{wang2023orthogonal} & \scalebox{0.90}{\no} & 73.32 & 68.37 & 50.26 & 61.12 & \underline{57.75} & 60.96 & 65.71 & 63.31 & 62.60  \\
& {\textbf{Octopus (ours)}} & \scalebox{0.90}{\no} & \textbf{81.73} & \textbf{84.73} & \textbf{52.30} & \textbf{62.83} & \textbf{57.77} & \underline{62.08} & \underline{67.36} & 64.72 & \textbf{66.69} \\

\hline\hline
\multirow{12}{*}{\rotatebox{90}{{Last}}}
& Sequential Finetune & - & 57.43 & 28.90 & 41.88 & 30.05 & 51.39 & 50.76 & 53.28 & 64.78 & 47.31  \\

& \multicolumn{11}{>{\columncolor{gray!30}}c}{\textcolor{olive!20!black}{$\triangledown$ Architecture-based}} \\
& L2P~\cite{wang2022learning} & \scalebox{0.90}{\no} & 70.21 & 23.31 & 44.21 & 43.76 & 56.25 & 58.46 & 62.32 & 64.11 & 52.83  \\
& MoELoRA~\cite{chen2024coin} & \scalebox{0.90}{\no} & 62.02 & 37.21 & 43.32 & 33.22 & 52.05 & 53.12 & 57.92 & \textbf{65.75} & 50.58  \\
& HiDe-LLaVA~\cite{guo2025hide} & \scalebox{0.90}{\no} & \underline{73.20} & \underline{69.28} & \underline{50.76} & \underline{59.18} & \underline{56.92} & \textbf{61.33} & \textbf{67.12} & 64.76 & \underline{62.81} \\

& \multicolumn{11}{>{\columncolor{gray!30}}c}{\textcolor{olive!20!black}{$\triangledown$ Regularization-based}} \\
& LwF~\cite{li2017learning} & \scalebox{0.90}{\no} & 60.71 & 30.58 & 41.49 & 36.01 & 52.80 & 47.07 & 53.43 & \underline{65.12} & 48.40  \\
& EWC~\cite{kirkpatrick2017overcoming} & \scalebox{0.90}{\no} & 59.75 & 31.88 & 42.26 & 34.96 & 51.06 & 51.84 & 55.30 & 64.55 & 48.95  \\
& O-LoRA~\cite{wang2023orthogonal} & \scalebox{0.90}{\no} & 72.56 & 62.84 & 48.43 & 58.97 & \textbf{57.66} & 59.14 & 63.21 & 63.31 & 60.77  \\
& {\textbf{Octopus (ours)}} & \scalebox{0.90}{\no} & \textbf{79.72} & \textbf{82.18} & \textbf{51.49} & \textbf{62.17} & 53.60 & \underline{60.50} & \underline{66.38} & 64.72 & \textbf{65.10}
 \\

\bottomrule
\end{tabular}
}
\end{center}
\vspace{-0.5cm}
\caption{Comparison with various methods on CoIN~\cite{chen2024coin} in terms of \emph{Avg} and \emph{Last}. The best and second methods are labeled with {bold} and {underline} styles. \textit{Zero-shot} evaluates the pretrained model without task-specific finetuning. \textit{Multi-task} jointly finetunes the model across all datasets, whereas \textit{Sequential Finetune} adapts only one LoRA module sequentially to individual tasks. These settings provide an empirical characterization of the lower bound, upper bound, and baseline for continual learning methods.}
\label{tab_coin}
\vspace{-0.2cm}
\end{table*}
\renewcommand{\arraystretch}{1.0}